\documentclass[journal,comsoc]{IEEEtran}
\usepackage{longtable}
\usepackage{times}
\usepackage{epsfig}
\usepackage{graphicx}
\usepackage{amsmath}
\usepackage{amssymb}
\usepackage{epsfig}
\usepackage{makecell,multirow,diagbox}
\usepackage[normalem]{ulem}
\usepackage{ulem}
\usepackage{comment}
\usepackage[colorlinks,linkcolor=red]{hyperref}
\usepackage{footnote}
\usepackage{threeparttable}
\usepackage{subfigure}

\usepackage{xcolor,soul,framed} 
\usepackage{array}
\usepackage{mdwmath}
\usepackage{mdwtab}
\usepackage{booktabs}
\usepackage[top=2cm, bottom=2cm, left=2cm, right=2cm]{geometry}
\usepackage{algorithm}
\usepackage{algpseudocode}

\ifCLASSINFOpdf
\else
\fi
\usepackage{amsmath}
\usepackage[cmintegrals]{newtxmath}
\makeatletter
\renewcommand{\@thesubfigure}{\hskip\subfiglabelskip}
 \makeatother

\begin{document}
%
\title{Learning from Pixel-Level Label Noise: A New Perspective for Semi-Supervised Semantic Segmentation}
%
%
%

\author{Rumeng~Yi,
        Yaping~Huang,
        Qingji~Guan,
        Mengyang~Pu,
        and~Runsheng~Zhang
\thanks{Rumeng Yi, Yaping Huang, Qingji Guan, Mengyang Pu and Runsheng Zhang are with the Beijing Key Laboratory of Traffic Data Analysis and Mining, Beijing Jiaotong University, Beijing 100044, China (e-mail: 19112038@bjtu.edu.cn; yphuang@bjtu.edu.cn; qingjiguan@gmail.com; mengyangpu@bjtu.edu.cn; rszhang@bjtu.edu.cn).}
}

%
%

\markboth{}%
{ Learning from Pixel-Level Label Noise: A New Perspective for Semi-Supervised Semantic Segmentation}
%



\maketitle

\begin{abstract}
   This paper addresses semi-supervised semantic segmentation by exploiting a small set of images with pixel-level annotations (strong supervisions) and a large set of images  with only image-level annotations (weak supervisions). Most existing approaches aim to generate accurate pixel-level labels from weak supervisions. However, we observe that those generated labels still inevitably contain noisy labels. Motivated by this observation, we present a novel perspective and formulate this task as a problem of learning with pixel-level label noise. Existing noisy label methods, nevertheless, mainly aim at image-level tasks, which can not capture the relationship between neighboring labels in one image. Therefore, we propose a graph based label noise detection and correction framework to deal with pixel-level noisy labels. In particular, for the generated pixel-level noisy labels from weak supervisions by Class Activation Map (CAM), we train a clean segmentation model with strong supervisions to detect the clean labels from these noisy labels according to the cross-entropy loss. Then, we adopt a superpixel-based graph to represent the relations of spatial adjacency and semantic similarity between pixels in one image. Finally we correct the noisy labels using a Graph Attention Network (GAT) supervised  by detected clean labels. We comprehensively conduct experiments on PASCAL VOC 2012, PASCAL-Context and MS-COCO datasets. The experimental results show that our proposed semi-supervised method achieves the state-of-the-art performances and even outperforms the fully-supervised models on PASCAL VOC 2012 and MS-COCO datasets in some cases.
\end{abstract}


%
\IEEEpeerreviewmaketitle

\section{Introduction}
%
%
%
%
\IEEEPARstart{B}{uilding} a large image dataset with high quality annotations for semantic segmentation is costly and time consuming. In order to tackle this problem, semi-supervised approaches have attracted more interests in recent years. Instead of pixel-level annotations, those methods make use of a small set of strong annotations and a large set of weak annotations, such as scribbles~\cite{lin2016scribblesup}~\cite{vernaza2017learning}, bounding boxes~\cite{khoreva2017simple} or image-level class labels~\cite{ahn2018learning}~\cite{wei2018revisiting}, to reduce the data annotation requirements. Among these three types of weak annotations, image-level class labels are the easiest one to obtain and have been widely used to bridge the gap between image-level and pixel-level annotations. In this paper, we address semi-supervised semantic segmentation using a small set of pixel-level strong annotations and a large set of image-level weak annotations.

\begin{figure}[t]
\centering
\subfigure[(a) Image]{
\begin{minipage}[t]{0.296\linewidth}
\centering
\includegraphics[width=1in]{orig_2007_001825.png}
\end{minipage}
}
\subfigure[(b) Ground truth]{
\begin{minipage}[t]{0.296\linewidth}
\centering
\includegraphics[width=1in]{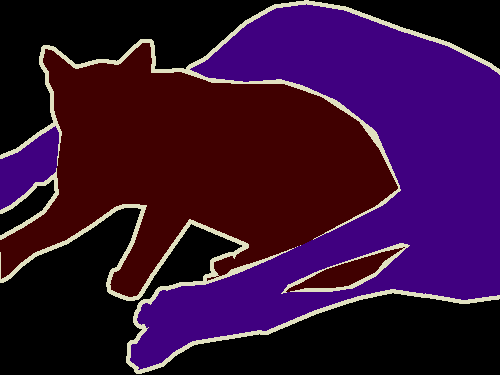}
\end{minipage}
}
\subfigure[(c) CAM]{
\begin{minipage}[t]{0.296\linewidth}
\centering
\includegraphics[width=1in]{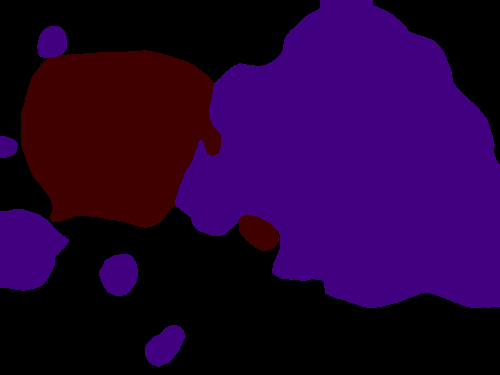}
\end{minipage}
}

\subfigure[(d) Initial seed regions
]{
\begin{minipage}[t]{0.295\linewidth}
\centering
\includegraphics[width=1in]{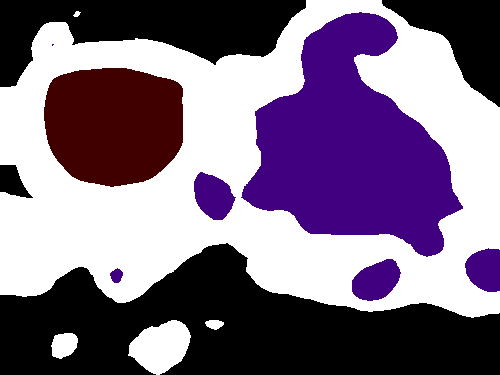}
\end{minipage}
}
\subfigure[(e) Our seed regions
]{
\begin{minipage}[t]{0.295\linewidth}
\centering
\includegraphics[width=1in]{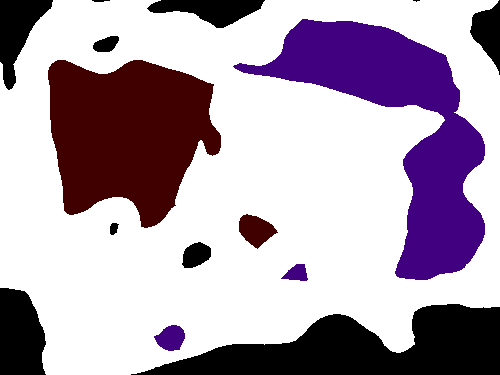}
\end{minipage}
}
\subfigure[(f) Our final prediction
]{
\begin{minipage}[t]{0.295\linewidth}
\centering
\includegraphics[width=1in]{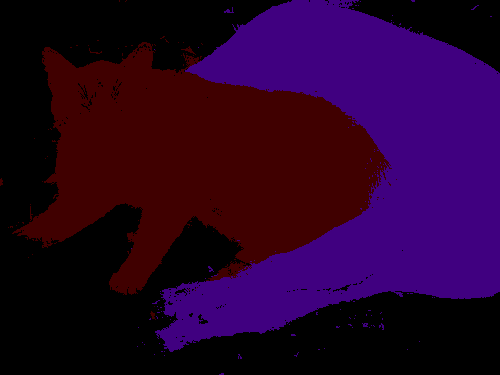}
\end{minipage}
}
\caption{For (a) an image with class labels, (b) ground truth, and (c) the original CAM, we choose the pixels with activation scores larger than 0.3 as foreground pixels, and the pixels smaller than 0.05 as background pixels to generate (d) initial seed regions (the black represents background and the white represents unlabeled pixels). However, we observe that these regions with high confidence contain many mislabeled pixels. Motivated by this, we propose a new perspective to generate (e) seed regions by detecting noisy labels. Subsequently, we design a graph based method to achieve better (f) pseudo labels by further correcting noisy labels.}

\label{Accuracy}
\end{figure}

The key step of semi-supervised methods is to infer accurate pixel-level labels for a large number of images with only image-level weak annotations. One popular choice is Class Activation Map (CAM)~\cite{zhou2016learning}, which highlights local discriminative regions by investigating the contribution of hidden units from a deep classification model. However, the generated object regions are small or sparse, and not sufficient to cover entire object area. In order to generate dense object regions, most previous methods regard these initial discriminative regions with high confidence as segmentation seeds. Then they focus on expanding them with different techniques like learning pixel-level affinities~\cite{ahn2018learning}, adversarial erasing manner~\cite{wei2017object} or stochastic regularization~\cite{lee2019ficklenet}. As shown in Figure~\ref{Accuracy}(d), we can see that the regions with high confidence do not mean that their corresponding labels are always correct. We experiment on the PASCAL VOC 2012 \emph{trainaug} set and the accuracy of seed regions is only \textbf{86.7\%}~\cite{ahn2019weakly}, which largely affect the following expansion process.

To tackle the problem, this paper proposes a novel perspective by observing the fact that the pixel-level labels generated by CAM contain a large number of noisy labels (compared with ground truth human annotations). 
We believe that semi-supervised semantic segmentation can be formulated as a problem of learning with pixel-level label noise. However, existing approaches of learning from noisy labels mainly aim at image classification tasks, such as CIFAR-100~\cite{krizhevsky2009learning}, Clothing-1M~\cite{xiao2015learning}, how to learn with pixel-level noisy labels has not been explored. Compared with image-level noisy labels, the critical issue of pixel-level noisy labels is to model the relations between pixel labels in one image. Therefore, we propose a graph based label noise detection and correction framework to capture the relations between neighboring pixels and further use these relations to correct the noisy labels. 

Specifically, we first adopt CAM to generate the pixel-level labels for the images with image-level labels. Then, we train a clean segmentation model with a small set of strong annotations to distinguish the clean labels from the noisy pixel-level labels according to the cross-entropy loss. A common method is to consider samples with smaller loss as clean ones~\cite{han2018co}~\cite{huang2019o2u}~\cite{li2020dividemix}. Subsequently, we construct a superpixel-based graph on one image by considering the dual constraints of spatial adjacency and semantic similarity, and thus we can correct noisy labels by embedding the clean labels into the graph and propagating the clean labels to the noisy labels using a Graph Attention Network (GAT)~\cite{velickovic2018graph}. Finally, the corrected pixel-level pseudo labels are used to train a semantic segmentation model. Compared with initial seed regions, the accuracy of our seed regions is largely improved (\textbf{96.7\%} vs \textbf{86.7\%} in PASCAL VOC 2012 \emph{trainaug} set). 

In summary, our contribution of this paper is three-fold:
\begin{itemize}
  \item We address semi-supervised semantic segmentation from a new perspective and formulate it as a problem of learning with pixel-level label noise.
   
  \item We propose a graph based label noise detection and correction framework to  deal with pixel-level noisy labels, which can  capture the relations between pixel-level labels in one image and correct the noisy labels efficiently.
  
  \item We conduct comprehensive experiments on the PASCAL VOC 2012, PASCAL-Context and MS-COCO datasets, and achieve the state-of-the-art performance. Especially, our proposed semi-supervised method even outperforms the fully-supervised models both on the PASCAL VOC 2012 and MS-COCO datasets in some cases.
\end{itemize}

\section{Related works}
\subsection{Semi-supervised semantic segmentation}
To reduce annotation effort, most existing approaches rely on semi-supervised training schemes, which use a small set of strong annotations and a large set of weak annotations, such as scribbles~\cite{lin2016scribblesup}~\cite{vernaza2017learning}, bounding boxes~\cite{khoreva2017simple} or image-level class labels~\cite{ahn2018learning}~\cite{wei2018revisiting}~\cite{huang2018weakly}. Among three types of weak annotations, image-level class labels are the easiest one to obtain. Class Activation Map (CAM)~\cite{zhou2016learning} is a good choice for generating pixel-level labels from image-level annotations, which can roughly localize object areas by drawing attentions on discriminative regions of class-specific objects. However, such localization maps are coarse and only activate parts of target objects. To address this issue, several techniques have been proposed to expand these activated regions to the whole target object~\cite{kolesnikov2016seed}~\cite{li2018tell}\cite{papandreou1502weakly}~\cite{wei2018revisiting}~\cite{lee2019ficklenet}. However, most of those methods need to carefully choose seed regions, which are only based on confidence maps produced by CAM. So they do not make full use of pixel-level strong supervisions when generating pseudo labels from weak annotations. Recently, a few studies provide novel perspectives to solve this problem. \cite{10.1007/978-3-030-58558-7_46} designs a strong-weak dual-branch network to exploit strong and weak annotations. \cite{Ouali_2020_CVPR} proposes a consistency training to enforce an invariance of the model's predictions over small perturbations applied to the inputs.

Different from the aforementioned approaches, we address semi-supervised semantic segmentation from a new perspective and formulate it as a pixel-level label noise learning problem. Therefore, we can generate more accurate pseudo labels from image-level annotations by making full use of a small set with pixel-level annotations.

\subsection{Learning with noisy labels}
Most large-scale datasets contain noisy labels. The techniques to alleviate the effect of noisy labels can be divided into two categories: (1) detecting noisy labels and then cleansing potential noisy labels or reduce their impacts~\cite{ren2018learning}~\cite{han2018co}~\cite{berthelot2019mixmatch}; (2) directly training noise-robust models with noisy labels~\cite{xiao2015learning}~\cite{guo2018curriculumnet}~\cite{li2019learning}.

So far, most existing methods are proposed to deal with image classification tasks. Only few works about noisy labels are applied to saliency detection~\cite{10.1007/978-3-030-58520-4_21}, anomaly detection~\cite{zhong2019graph} or instance segmentation~\cite{yang2020lncis}. In this paper, we propose the first usage of learning with noisy labels for semi-supervised semantic segmentation task, which can be considered as a pixel-wise classification problem. However, relations between the pixel labels need to be adequately modeled, and very few studies have explicitly addressed this with unreliable and noisy labels. Inspired by the previous researches, we propose a graph based label noise detection and correction framework and attempt to learn with noisy labels in pixel-level segmentation task.

\subsection{Dealing with graph-structured data}
Generalizing CNNs to inputs with graph-structured data is an important topic in the field of deep learning. Advances in this direction are often categorized as spectral-based approaches and spatial-based approaches. Spectral-based approaches work with a spectral representation of the graphs and have been successfully applied in the context of node classification. The first prominent spectral research on spectral Graph Convolutional Network (GCN) is presented in~\cite{bruna2013spectral}. Later, subsequent studies aim to designing more efficient and flexible spectral-based solutions~\cite{kipf2016semi}~\cite{defferrard2016convolutional}~\cite{henaff2015deep}. In all of the aforementioned spectral approaches, the learned filters depend on the Laplacian eigenbasis. Spatial-based approaches define graph convolutions based on a node's spatial relations, where an image can be considered as a special form of a graph with each pixel representing a node~\cite{wu2020comprehensive}. Graph Attention Network (GAT)~\cite{velickovic2018graph} is a representative approach toward spatial-based methods. It introduces an attention\textrm{-}based architecture to perform node classification of graph-structured data, which computes the hidden representations of each node in the graph by attending over its neighbors.

\begin{figure*}[t]
\begin{center}
\includegraphics[width=0.921\linewidth]{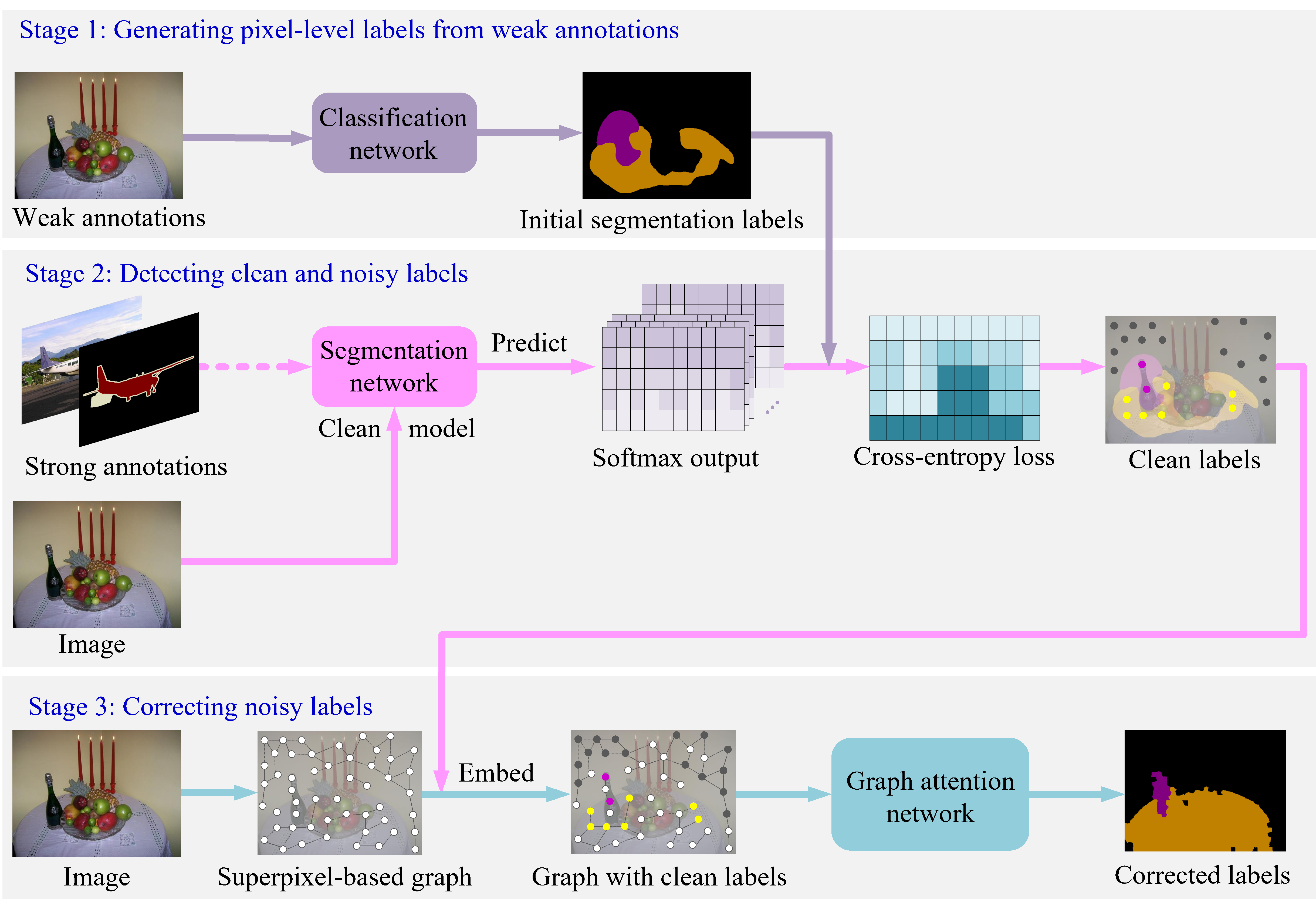}
\caption{Overview of the label noise detection and correction framework for semi-supervised semantic segmentation. Firstly, we adopt CAM to generate the pixel-level labels for a large set of images with weak annotations as the initial segmentation labels. Secondly, we train a clean segmentation model with strong annotations to detect clean labels from the initial segmentation labels. Then we construct a superpixel-based graph on one image and embed the clean labels into the graph. Finally, we correct the noisy labels by using a GAT.
}
\label{framework}
\end{center}
\end{figure*}

Recently, many studies attempt to apply these techniques in computer vision tasks, such as scribble and bounding box-based weakly semantic segmentation~\cite{pu2018graphnet}, image-text matching~\cite{li2019visual}, visual question answering~\cite{li2019relation}, semantic object parsing~\cite{liang2016semantic} and human-object interaction detection~\cite{linaction}. The most relevant work is~\cite{pu2018graphnet}, which uses GCN to deal with scribble and bounding box-based weakly semantic segmentation. However, our approach is fundamentally different from~\cite{pu2018graphnet}. Directly extending~\cite{pu2018graphnet} does not deal with image-level annotations. The reason is that GCN heavily relies on the supervision quality. But the generated labels by CAM contain a large number of noise compared with scribble and bounding box, which leads to poor performance. Therefore, one key contribution of our work is to detect clean labels from the perspective of learning with noisy labels, which is the crucial foundation of seed-and-expansion strategies.

\section{Proposed method}
In this section, we give the details of our proposed label noise detection and correction framework for semi-supervised semantic segmentation. Figure~\ref{framework} shows the overall framework. Firstly, we adopt CAM to generate the pixel-level labels from image-level annotations and regard them as initial segmentation labels, which contain a large number of noisy labels. Secondly, in order to ensure that the clean labels are detected from the initial segmentation labels accurately, we train a clean segmentation model by using a small set of strong annotations to guide the detection process. Then, we construct a superpixel-based graph on one image by considering the dual constraints of spatial adjacency and semantic similarity, and adopt GAT to correct noisy labels supervised by clean labels. Finally, the corrected labels are used as pseudo labels to train a segmentation network in a semi-supervised manner.

\subsection{Generating pixel-level labels from weak annotations}

We follow the approach of~\cite{zhou2016learning} to compute CAMs of training images as initial segmentation labels. The architecture is a typical classification network with global average pooling (GAP) followed by a fully connected layer, which is trained with image-level labels. Given an image, the CAM of a ground truth class $c$ is computed by:
\begin{equation}
  {{M}_{c}}(x,y)=\sum\limits_{k}{\omega _{k}^{c}{{f}_{k}}(x,y)}
\end{equation}

where ${{M}_{c}}$ is the class activation map for class $c$, ${{f}_{k}}(x,y)$ is the activations of unit $k$ in the last convolutional layer at spatial location $(x,y)$, and $\omega _{k}^{c}$ is the weight corresponding to class $c$ for unit $k$. Besides, for any class $c'$ irrelevant to the ground truths, we disregard ${{M}_{{{c}^{'}}}}$ by setting its activation scores to zero. Thus we can assign the corresponding class label to each pixel according to the highest activation score, and the pixels with activation scores smaller than 0.05 are background. These generated pixel-level labels can be used as the initial segmentation labels, which obviously contain inevitable noisy labels.

\subsection{Detecting clean and noisy labels}
The key issue of our proposed method is to detect clean and noisy labels from the initial segmentation labels computed by CAM. In this section, we propose to use a small set of pixel-level annotations to train a clean segmentation model, and then detect the clean and noisy labels according to the cross-entropy loss of generated pixel-level labels trained on the clean model.

Specifically, our training dataset $\mathcal{D}$ comprises of two subsets: ${{\mathcal{D}}^{c}}:\{(x_{i}^{c},y_{i}^{c}),1\le i\le M\}$ with a small set of pixel-level annotations, where ${{y}_{i}^{c}}$ is the ground truth pixel-level labels corresponding to $i\textrm{-}th$ input image $x_{i}^{c}$; and ${{\mathcal{D}}^{n}}:\{(x_{j}^{n},y_{j}^{n}),1\le j\le N, M\ll N\}$ with image-level annotations, where ${{x}_{j}^{n}}$ is the $j\textrm{-}th$ input image and ${{y}_{j}^{n}}$ is the corresponding pixel-level labels produced by CAM. Firstly, we train a segmentation network on ${{\mathcal{D}}^{c}}$ and obtain a clean segmentation model $\mathcal{C}$. Then we use $\mathcal{C}$ to predict the label for each pixel $p$ of $x_{j}^{n}$ in ${{\mathcal{D}}^{n}}$ and calculate their cross-entropy loss ${{{l}}_{jp}}$ under the supervision of $y_{jp}^{n}$: 
\begin{equation}
{{l}_{jp}}=-y_{jp}^{n}\log ({{F}_{\mathcal{C}}}(x_{jp}^{n}))
\end{equation}
where ${{F}_{\mathcal{C}}}$ is a segmentation function of clean model $\mathcal{C}$ that projects the image to the prediction. Following many noisy label learning approaches which consider samples with smaller loss as clean ones~\cite{han2018co}~\cite{arazo2019unsupervised}~\cite{ren2018learning}, we set a threshold $\theta$ to distinguish the labels with small loss as the clean labels  and then use these clean labels as the supervisory information for training GAT.

Because the threshold $\theta$ plays an important role in detecting clean and noisy labels. In this paper, we introduce an efficient strategy to select a suitable threshold. At first, we generate the  pixel-level labels by CAM on ${{\mathcal{D}}^{c}}$. Then we use the clean model $\mathcal{C}$ to predict the label for each pixel in ${{\mathcal{D}}^{c}}$ and calculate their cross-entropy loss under the supervision of pixel-level labels computed by CAM. Since we have the ground truth labels on ${{\mathcal{D}}^{c}}$, we can easily choose a suitable threshold according to the loss distributions of the clean labels. It should be noted that we need to consider both the quantity and quality of the labels we choose, which means that we need to guarantee that not only the labels we choose are correct as much as possible but also the number of selected labels is not too little. The detailed experiments will be given in ablation study.

\subsection{Correcting noisy labels}
In this section, we aim to correct the pixel-level noisy labels which are selected in detecting clean and noisy labels stage. However, most existing approaches of learning with noisy labels mainly focus on image classification tasks, which do not consider to model the relations of the pixel labels in one image. Therefore, we develop a graph attention network-based method to correct the pixel-level noisy labels. Specifically, we first construct a superpixel-based graph on an image by considering the dual constraints of \textit{spatial adjacency} and \textit{semantic similarity}~\cite{pu2018graphnet}. Then the clean labels are embedded into the graph. Finally, we employ GAT to propagate the label information from clean labels to noisy labels.

It should be noted that other message-passing neural networks that operate on graph-structured data can be also employed to perform the noisy labels correction in our framework, such as GCN~\cite{kipf2016semi}. Considering the better performance, we apply GAT in our paper. The comparison between GCN and GAT is given in ablation study.

\subsubsection{Superpixel-based graph construction}

Considering that superpixel can provide a larger, locally homogeneous and coherent regions that preserve most of the structure necessary for accurate segmentation~\cite{vieux2012segmentation}, we transform an image to a superpixel-based graph representation $\mathcal{G}=(\mathcal{V},\mathcal{E},\mathcal{A})$, where $\mathcal{V}$ is a set of nodes, $\mathcal{E}$ is a set of edges and $\mathcal{A}$ is an adjacency matrix. In particular, $\mathcal{V}$ denotes the superpixel set $\{s{{p}_{i}}\}_{i=1}^{N}$, the edge ${{\varepsilon }_{ij}}$ in $\mathcal{E}$ connects $s{{p}_{i}}$ and $s{{p}_{j}}$ where the two nodes are neighboring in space, and $\mathcal{A}$ represents the nodes proximity. 

\textbf{Vertex construction.} Firstly, we use the Simple Linear Iterative Clustering (SLIC) \cite{achanta2012slic} method to over-segment one image and divide it into a superpixel set, denoted as $\{s{{p}_{i}}\}_{i=1}^{N}$, where $N$ is the number of superpixels contained in one image. In our experiments, one image is over-segment into about 1000 superpixels. Then we use the conv5-3 layer of the clean model $\mathcal{C}$ to extract the high-level semantic features from the whole image and integrate over superpixels. Since the feature maps are downsampled by the operation of several convolutional layers, we apply a bilinear interpolation on the feature maps to obtain the dense feature maps with the same size as the original images, and then an average pooling is performed on a superpixel along the channels. Finally, a 512-dimensional CNN feature vector for each superpixel $s{{p}_{i}}$ is obtained.

\textbf{Edge construction.} We leverage two characteristics of images to construct edges of the graph on an image, \textit{i.e., spatial adjacency} and \textit{semantic similarity}. Intuitively, spatial adjacency means neighboring pixels tend to have similar labels, while semantic similarity means the pixels with the same labels probably share similar semantic information. We assume that two spatially adjacent nodes which have similar semantic content commonly tend to belong to the same class. Thus, we consider the dual constraints of spatial adjacency and semantic similarity to construct edges.

Firstly, we model the spatial adjacency on $\mathcal{G}$. We construct the \emph{spatial adjacency weight matrix} \begin{small}$ {{W}_{l}}={{[ w_{l}^{ij}]}_{n\times n}}\in {{\mathbb{R}}^{N\times N}}$ \end{small} to measure the spatial adjacent relationship between $s{{p}_{i}}$ and all other superpixels. If $s{{p}_{i}}$ and $s{{p}_{j}}$ are spatially adjacent, then the weight \begin{small}$w_{l}^{ij}$\end{small} is set to 1, otherwise \begin{small}$w_{l}^{ij}$\end{small} is set to 0. Then we model the semantic similarity on $\mathcal{G}$. Specifically, we construct a \textit{semantic similarity weight matrix} \begin{small}${{W}_{s}}={{[ w_{s}^{ij}]}_{n\times n}}\in {{\mathbb{R}}^{N\times N}}$\end{small} to measure the semantic similarity between $s{{p}_{i}}$ and its spatially neighboring superpixels. Given the 512-dimensional feature vector $\{ {{v}_{i}} \}_{i=1}^{N}$ for each superpixel $s{{p}_{i}}$, the weight $w_{s}^{ij}$ is given by:
\begin{equation}
w_{s}^{ij}=w_{l}^{ij}\times \exp (-\frac{\left\| {{v}_{i}}-{{v}_{j}} \right\|}{2h})
\end{equation}
where $h$ is the dimension of the feature vector. If two superpixels $s{{p}_{i}}$ and $s{{p}_{j}}$ are not spatially adjacent, $i.e.,$ $w_{l}^{ij}=0$, we will ignore their semantic relationship.

Subsequently, the adjacency matrix $\mathcal{A}={{\left[ {{a}_{ij}} \right]}_{n\times n}}\in {{\mathbb{R}}^{N\times N}}$ is given by:

\begin{equation}
{{a}_{ij}}=\left\{ 
            \begin{aligned}
              & 0\text{, \; if\; }w_{s}^{ij}<\gamma \text{\;}\!\!\And\!\!\text{  }w_{s}^{ij}<\underset{k\in {{\mathcal{N}}_{\mathcal{G}}}(i)}{\mathop{\max }}\,(w_{s}^{ik}) \\  
            & 1\text{, \; otherwise} \\ 
            \end{aligned}
            \right.
\label{adj matrix}
\end{equation}
where $\gamma $ is a threshold to filter out the edges with low similarity from the edge set $\mathcal{E}$. In our work, $\gamma =u({{W}_{s}})-\sigma ({{W}_{s}})$, where $u(\cdot )$ and $\sigma (\cdot )$ are mean and standard deviation, respectively. $\underset{k\in {{\mathcal{N}}_{\mathcal{G}}}(i)}{\mathop{\max }}\,(w_{s}^{ik}) $ is the maximum semantic similarity between $s{{p}_{i}}$ and its neighboring superpixels $s{{p}_{k}}$. Moreover, if $w_{s}^{ij}$ is lower than both $\gamma $ and $\underset{k\in {{\mathcal{N}}_{\mathcal{G}}}(i)}{\mathop{\max }}\,(w_{s}^{ik}) $, the edge ${{\varepsilon }_{ij}}$ is removed.

\subsubsection{Correcting noisy labels}
After constructing the superpixel-based graph for each image, the clean labels which are selected in detecting clean and noisy labels stage are embedded into the graph as the supervision information for training GAT.

\textbf{Embedding clean labels.} We denote the clean labels set of each image as $\mathcal{S}=\{{{s}_{k}},{{c}_{k}}\}_{k=1}^{K}$, where ${{s}_{k}}$ is the $k\textrm{-}th$ pixel and ${{c}_{k}}$ is the category label of ${{s}_{k}}$. If $s{{p}_{i}}$ overlaps with ${{s}_{k}}$ ($s{{p}_{i}}\cap {{s}_{k}}\ne \emptyset $), we will assign the category label ${{c}_{k}}$ to $s{{p}_{i}}$. However, we find that one superpixel usually contains more than one clean label in our experiments, so we need to select the corresponding label with the largest number of clean labels and assign it to the superpixel.

\textbf{Correcting noisy labels by GAT.} We transform an image to a superpixel-based graph representation and leverage GAT~\cite{velickovic2018graph} to correct noisy labels. Following~\cite{velickovic2018graph}, we perform \textit{self\textrm{-}attention} on the nodes to compute attention coefficients. The importance of node \textit{j'}s features to node \textit{i} is given by:
\begin{equation}
{{e}_{ij}}=\varphi {{({{v}_{i}})}^{T}}\phi ({{v}_{j}})
\end{equation}
where $\varphi ({{v}_{i}})={{W}_{\varphi }}{{v}_{i}}$ and $\phi ({{v}_{j}})={{W}_{\phi }}{{v}_{j}}$. ${{W}_{\varphi }}$ and ${{W}_{\phi }}$ are the weight parameters learned by back propagation. We only compute ${{e}_{ij}}$ for nodes $j\in {{\mathcal{N}}_{i}}$, where ${{\mathcal{N}}_{i}}$ is the set of neighboring nodes of node \textit{{}i} in the graph. To make coefficients easily comparable across different nodes, we normalize ${{e}_{ij}}$ across all choices of \textit{j} using the softmax function:
\begin{equation}
{{\alpha }_{ij}}=\frac{\exp ({{e}_{ij}})}{\sum\nolimits_{k\in {{\mathcal{N}}_{i}}}{\exp ({{e}_{ik}})}}\
\end{equation}

Then we extend the above graph attention mechanism by employing multi-head attention, and the final output features for each node are given by:
\begin{equation}
    v_{i}^{'}=\parallel _{l=1}^{L}\sigma (\sum\limits_{j}{\alpha _{ij}^{l}W_{g}^{l}{{v}_{j}}})
\label{gat_feature}
\end{equation}
where $\parallel $ denotes concatenation, $\sigma (\cdot )$ is a nonlinear function such as ReLU, ${\alpha _{ij}^{l}}$ are normalized attention coefficients computed by the $l$-th attention mechanism, and ${W_{g}^{l}}$ is the corresponding input linear transformation's weight matrix.

In our experiment, we adopt a two-layer GAT for label correction. The forward model is given by:
\begin{equation}
{Z=f(V,A)}
\end{equation}
where $V$ is the superpixel-based feature matrix computed by Eq.\ref{gat_feature}, A is the adjacency matrix computed by Eq.\ref{adj matrix}, and the cross-entropy loss over all superpixels with clean labels is defined by:
\begin{equation}
\mathcal{L}=-\sum\limits_{i=1}^{P}{{{p}_{i}}\ln {{z}_{i}}}
\end{equation}
where ${p_{i}}$ is the corresponding label of $s{{p}_{i}}$, $P$ is the number of the superpixels with clean labels and $z_{i}$ is the GAT's prediction of $s{{p}_{i}}$.

\subsection{Training the segmentation network}
The category information of pixels which were originally considered to be noise in each image has been corrected by GAT, thus we can recover the label of each pixel according to their corresponding superpixel and then the corrected segmentation labels are refined by dense CRF \cite{krahenbuhl2011efficient} to better estimate object shapes.

The segmentation labels obtained by detecting and correcting stages are finally used as supervision to train a segmentation network. Note that any fully supervised semantic segmentation model can be employed in our approach.

\section{Experiments}
In this section, we evaluate our approach on PASCAL VOC 2012~\cite{everingham2015pascal}, PASCAL-Context~\cite{mottaghi2014role} and MS-COCO~\cite{lin2014microsoft} datasets, where our framework generates segmentation labels as ground truth labels for training a segmentation model. The performance is measured using the mean Intersection-over-Union (mIoU) across the available classes.

\textbf{Datasets.} We evaluate our approach on PASCAL VOC 2012, PASCAL-Context and MS-COCO datasets. The original PASCAL VOC 2012 dataset consists of 1464 training, 1449 validation, and 1456 test images covering 21 classes (one background class). An auxiliary dataset of 9118 training images is provided by~\cite{hariharan2011semantic}. PASCAL-Context dataset is a whole scene parsing dataset containing 4998 training and 5105 testing images with dense semantic labels, and this dataset involves 60 classes (one background class). MS-COCO dataset contains 81 classes (one background class), 80k and 40k images for training and validation.

\textbf{Implementation details.} We use the DeepLab-CRF-LargeFOV model~\cite{chen2014semantic} where the parameters are initialized by VGG-16~\cite{simonyan2014very} network pre-trained on the ImageNet dataset~\cite{russakovsky2015imagenet} (hereinafter referred to as "DeepLab v1-vgg16") and Deeplab v2 model~\cite{chen2017deeplab} where the parameters are initialized by resnet-101 network~\cite{7780459} pre-trained on the ImageNet dataset (hereinafter referred to as "DeepLab v2-resnet101") as the basic network. And we use the standard settings for all parameters.

\subsection{Results on PASCAL VOC 2012 dataset}
\label{section:Results on PASCAL VOC 2012 dataset}

\begin{table*}[]
\centering
\caption{\centering Per-class IoU on PASCAL VOC 2012 validation and test set with different ${{\mathcal{D}}^{c}}$ using Deeplab v1-vgg16 segmentation network.}
\setlength{\tabcolsep}{1.2mm}{
\begin{tabular}{lcccccccccccccccccccccc}
\hline
{${{\mathcal{D}}^{c}}$}   &{bkg}  &{aero}  &{bike}  &{bird}  &{boat}  &{bottle}  &{bus}  &{car}  &{cat}  &{chair}  &{cow}  &{table}  &{dog}  &{horse}  &{mbk}  &{person}  &{plant}  &{sheep} &{sofa} &{train} &{tv} &{mean} \\ \hline
\multicolumn{23}{l}{Results on validation set:}                                                    \\
200 &{90.0}  &{76.6}  &{29.2}  &{80.6}  &{55.4}  &{64.2}  &{80.0}  &{73.9}  &{80.5}  &{29.1}  &{70.6}  &{49.5}  &{75.2}  &{68.1}  &{65.8}  &{75.9}  &{41.0}  &{74.9} &{39.7} &{61.7} &{56.1} &{\textbf{63.7}}     \\
400   &{90.2}  &{76.6}  &{29.6}  &{80.2}  &{55.8}  &{64.8}  &{80.1}  &{74.3}  &{80.2}  &{29.2}  &{70.8}  &{49.4}  &{73.7}  &{68.0}  &{65.7}  &{76.4}  &{46.0}  &{72.9}  &{38.6} &{63.5} &{58.6} &{\textbf{64.0}}    \\
800 &{90.3}  &{76.3}  &{30.4}  &{80.8}  &{55.9}  &{66.8}  &{81.5}  &{73.6}  &{81.4}  &{30.9}  &{74.3}  &{49.3}  &{75.5}  &{69.6}  &{66.2}  &{77.1}  &{45.6}  &{74.4}  &{39.3} &{66.3} &{58.3} &{\textbf{64.9}}     \\
1464  &{92.1}  &{80.9}  &{32.6}  &{80.8}  &{67.3}  &{68.1}  &{83.3}  &{77.2}  &{79.4}  &{33.0}  &{71.5}  &{54.3}  &{74.0}  &{71.5}  &{69.8}  &{79.3}  &{44.7}  &{76.3}  &{45.6} &{74.6} &{62.1} &{\textbf{67.5}}   \\ \hline
\multicolumn{23}{l}{Results on test set:}                                                   \\
200  &{90.3}  &{74.4}  &{33.6}  &{81.3}  &{54.9}  &{60.4}  &{77.5}  &{74.0}  &{80.0}  &{26.0}  &{63.7}  &{53.9}  &{75.6}  &{68.6}  &{76.3}  &{74.7}  &{46.6}  &{76.9} &{43.8} &{61.6} &{54.2} &{\textbf{64.2}}    \\
400  &{90.7}  &{75.7}  &{33.1}  &{80.4}  &{55.8}  &{61.1}  &{77.0}  &{74.2}  &{82.8}  &{25.9}  &{64.5}  &{55.8}  &{76.2}  &{69.3}  &{76.4}  &{75.3}  &{49.4}  &{75.3}  &{45.9} &{65.0} &{55.5} &{\textbf{65.0}}    \\
800  &{90.8}  &{78.1}  &{32.9}  &{78.0}  &{56.1}  &{61.2}  &{78.0}  &{73.8}  &{82.4}  &{26.4}  &{68.6}  &{57.3}  &{76.9}  &{71.9}  &{76.3}  &{76.9}  &{45.6}  &{77.3}  &{48.3} &{64.6} &{56.0} &{\textbf{65.6}}    \\
1464 &{92.4}  &{78.9}  &{38.8}  &{78.5}  &{59.7}  &{63.7}  &{83.1}  &{78.4}  &{80.0}  &{27.1}  &{70.6}  &{59.9}  &{75.3}  &{72.1}  &{79.5}  &{77.9}  &{50.4}  &{82.1}  &{46.2} &{71.5} &{59.2} &{\textbf{67.9}}   \\ \hline
\end{tabular}}
\label{per-class voc v1}
\end{table*}

\begin{table*}[]
\centering
\caption{\centering Per-class IoU on PASCAL VOC 2012 validation and test set with different ${{\mathcal{D}}^{c}}$ using Deeplab v2-resnet101 segmentation network.}
\setlength{\tabcolsep}{1.2mm}{
\begin{tabular}{lcccccccccccccccccccccc}
\hline
{${{\mathcal{D}}^{c}}$}   &{bkg}  &{aero}  &{bike}  &{bird}  &{boat}  &{bottle}  &{bus}  &{car}  &{cat}  &{chair}  &{cow}  &{table}  &{dog}  &{horse}  &{mbk}  &{person}  &{plant}  &{sheep} &{sofa} &{train} &{tv} &{mean} \\ \hline
\multicolumn{23}{l}{Results on validation set:}                                                    \\
200 &{91.6}  &{83.0}  &{35.5}  &{79.5}  &{62.6}  &{71.8}  &{87.1}  &{80.9}  &{85.1}  &{33.9}  &{75.6}  &{54.6}  &{78.9}  &{72.9}  &{75.9}  &{82.8}  &{50.4}  &{72.5} &{40.7} &{76.2} &{67.3} &{\textbf{69.5}}     \\
400  &{91.8}  &{83.5}  &{36.1}  &{79.4}  &{62.8}  &{72.9}  &{88.2}  &{81.3}  &{86.6}  &{33.6}  &{78.5}  &{57.2}  &{80.5}  &{74.7}  &{72.5}  &{81.5}  &{52.1}  &{74.4} &{42.5} &{74.7} &{59.7} &{\textbf{69.7}}     \\
800  &{92.1}  &{84.1}  &{37.3}  &{80.5}  &{62.4}  &{72.8}  &{88.5}  &{80.0}  &{86.3}  &{31.9}  &{79.7}  &{59.7}  &{80.4}  &{79.2}  &{76.1}  &{82.4}  &{56.8}  &{80.1} &{43.8} &{78.5} &{64.8} &{\textbf{71.3}}    \\
1464   &{94.4}  &{89.0}  &{56.4}  &{84.9}  &{71.9}  &{76.5}  &{93.2}  &{85.6}  &{89.0}  &{35.6}  &{86.6}  &{62.8}  &{84.3}  &{85.9}  &{83.0}  &{86.4}  &{53.9}  &{86.6} &{49.5} &{87.9} &{75.8} &{\textbf{77.1}}    \\ \hline
\multicolumn{23}{l}{Results on test set:}                                                   \\
200 &{91.9}  &{85.0}  &{35.8}  &{84.1}  &{58.1}  &{67.1}  &{88.7}  &{79.2}  &{85.4}  &{34.1}  &{76.1}  &{58.2}  &{82.3}  &{78.1}  &{79.4}  &{78.7}  &{48.7}  &{78.4} &{56.5} &{70.8} &{56.0} &{\textbf{70.1}}      \\
400  &{92.0}  &{86.4}  &{36.3}  &{85.9}  &{57.6}  &{68.4}  &{87.6}  &{79.4}  &{82.5}  &{34.6}  &{73.7}  &{57.6}  &{81.0}  &{76.6}  &{82.6}  &{78.6}  &{43.2}  &{77.3} &{58.8} &{73.8} &{61.3} &{\textbf{70.2}}     \\
800   &{92.5}  &{85.7}  &{36.7}  &{79.9}  &{57.5}  &{67.6}  &{88.1}  &{79.2}  &{88.3}  &{35.8}  &{78.7}  &{62.3}  &{83.0}  &{80.6}  &{83.1}  &{80.0}  &{54.1}  &{80.4} &{59.8} &{73.2} &{61.4} &{\textbf{71.8}}     \\
1464 &{94.6}  &{90.4}  &{52.2}  &{87.7}  &{61.8}  &{74.8}  &{93.3}  &{84.9}  &{90.0}  &{35.3}  &{85.1}  &{66.4}  &{87.9}  &{88.1}  &{84.4}  &{85.3}  &{56.8}  &{88.1} &{58.7} &{84.0} &{70.7} &{\textbf{77.2}}    \\ \hline
\end{tabular}}
\label{per-class voc v2}
\end{table*}

\textbf{Experimental results.} We compare our approach with current state-of-the-art methods~\cite{wei2018revisiting} \cite{lee2019ficklenet} \cite{huang2018weakly} \cite{li2018tell} \cite{papandreou1502weakly}  \cite{Ouali_2020_CVPR} \cite{10.1007/978-3-030-58558-7_46}. We adopt the same strong/weak split, $i.e.,$ 1.4K strongly annotated images and 9K weakly annotated images. Table~\ref{Comparison} summarizes the comparison. To the best of our knowledge, most  semi-supervised methods report the results on Deeplab v1-vgg16, only \cite{Ouali_2020_CVPR} and \cite{10.1007/978-3-030-58558-7_46} are performed on resnet50 \cite{Ouali_2020_CVPR} and  Deeplab v2-resnet101 \cite{10.1007/978-3-030-58558-7_46}. Therefore, for a fair comparison, we illustrate the backbone used for each approach in Table~\ref{Comparison}, and evaluate our method on different backbones. From the results, we can see that our proposed method outperforms other methods and achieves the best mIoU of 67.5\% and 67.9\% (v1-vgg16) and 77.1\% and 77.2\% (v2-resnet101) on \emph{val} and test sets, respectively. It should be noted that the performance on \emph{val} set  outperforms the fully-supervised model (76.5\%) by 0.6\% using v2-resnet101 segmentation network. We explain this surprising result and find a possible explanation. We believe the pseudo labels corrected by GAT can successfully highlight some instances which are improperly labeled in ground truth annotations, thus resulting in an improvement in the performance of the segmentation network, which exceeds the fully-supervised model. Moreover, our proposed method dedicates to generate more accurate pseudo-labels, and thus the performance can be further improved by incorporating the strong-weak dual-branch network~\cite{10.1007/978-3-030-58558-7_46}.

\begin{figure*}[htbp]
\centering
\subfigure{
\begin{minipage}[t]{0.078\linewidth}
\includegraphics[width=0.68in]{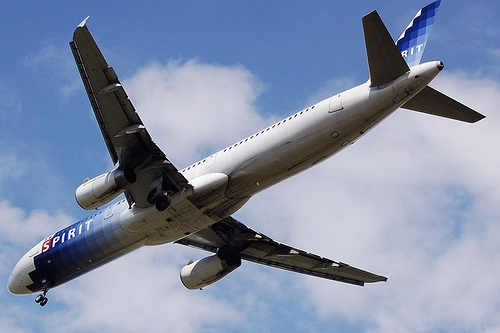}
\end{minipage}
}
\subfigure{
\begin{minipage}[t]{0.078\linewidth}
\includegraphics[width=0.68in]{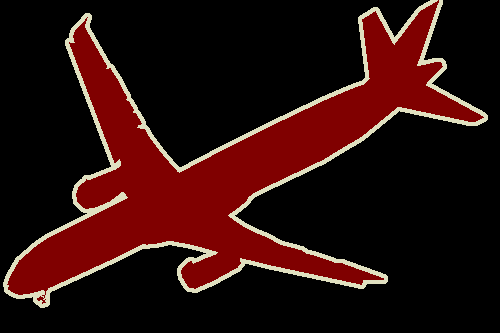}
\end{minipage}
}
\subfigure{
\begin{minipage}[t]{0.078\linewidth}
\includegraphics[width=0.68in]{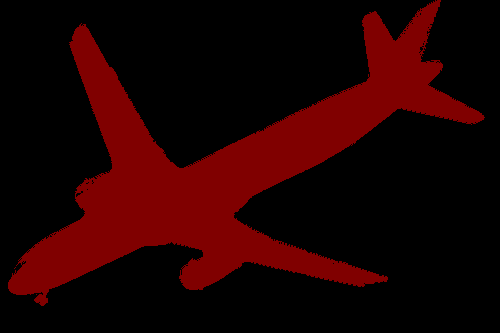}
\end{minipage}
}
\subfigure{
\begin{minipage}[t]{0.078\linewidth}
\includegraphics[width=0.68in]{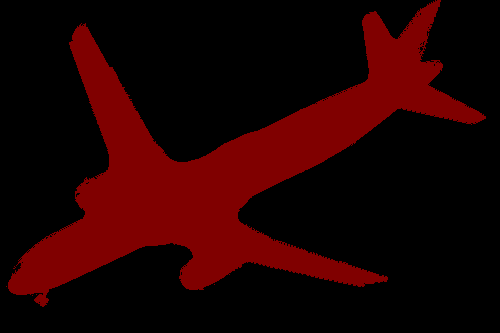}
\end{minipage}
}
\subfigure{
\begin{minipage}[t]{0.078\linewidth}
\includegraphics[width=0.68in]{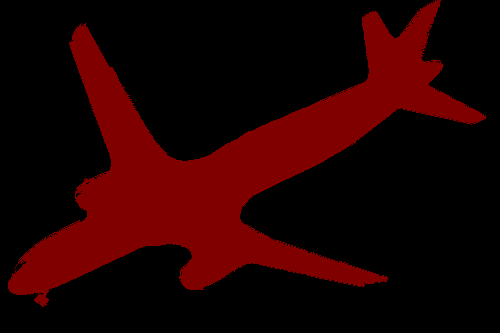}
\end{minipage}
}
\subfigure{
\begin{minipage}[t]{0.078\linewidth}
\includegraphics[width=0.68in]{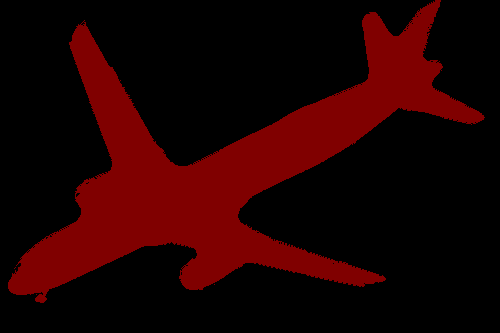}
\end{minipage}
}
\subfigure{
\begin{minipage}[t]{0.078\linewidth}
\includegraphics[width=0.68in]{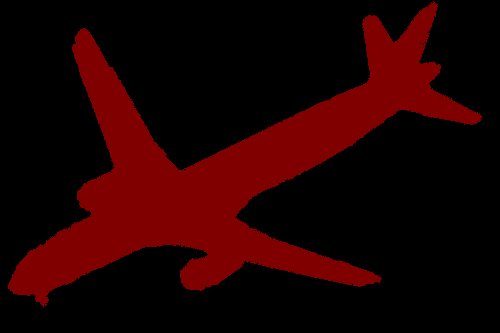}
\end{minipage}
}
\subfigure{
\begin{minipage}[t]{0.078\linewidth}
\includegraphics[width=0.68in]{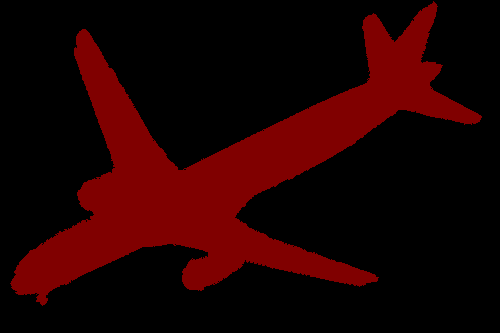}
\end{minipage}
}
\subfigure{
\begin{minipage}[t]{0.078\linewidth}
\includegraphics[width=0.68in]{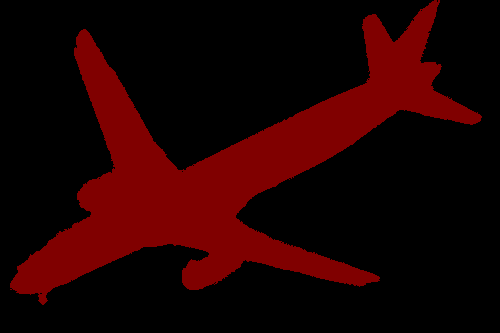}
\end{minipage}
}
\subfigure{
\begin{minipage}[t]{0.078\linewidth}
\includegraphics[width=0.68in]{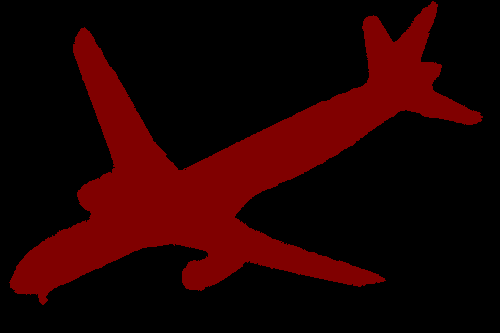}
\end{minipage}
}
\vspace{-8pt}

\subfigure{
\begin{minipage}[t]{0.078\linewidth}
\includegraphics[width=0.68in]{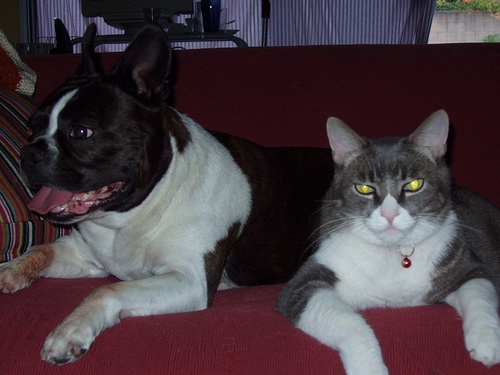}
\end{minipage}
}
\subfigure{
\begin{minipage}[t]{0.078\linewidth}
\includegraphics[width=0.68in]{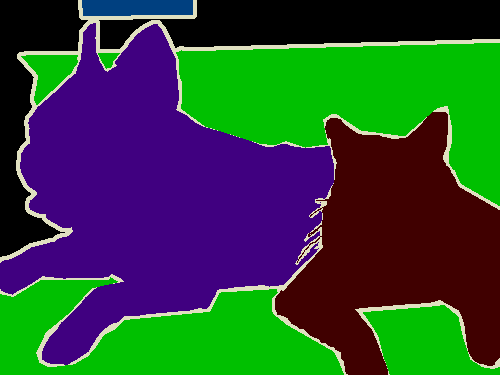}
\end{minipage}
}
\subfigure{
\begin{minipage}[t]{0.078\linewidth}
\includegraphics[width=0.68in]{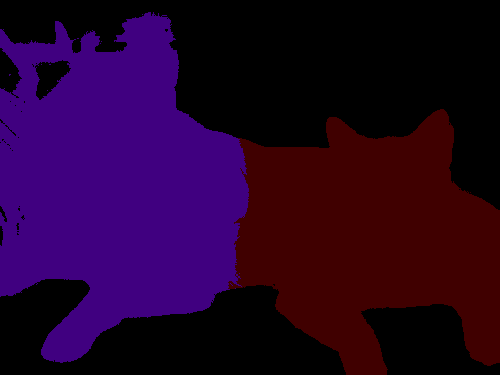}
\end{minipage}
}
\subfigure{
\begin{minipage}[t]{0.078\linewidth}
\includegraphics[width=0.68in]{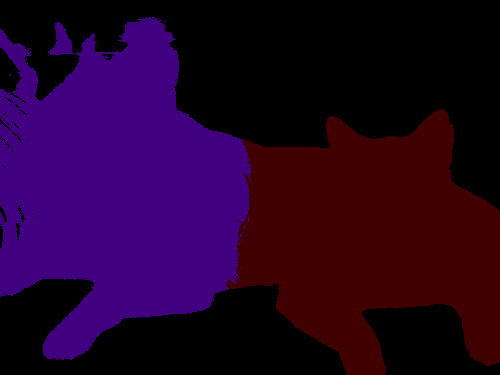}
\end{minipage}
}
\subfigure{
\begin{minipage}[t]{0.078\linewidth}
\includegraphics[width=0.68in]{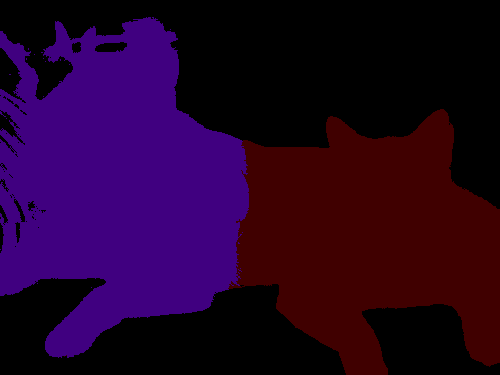}
\end{minipage}
}
\subfigure{
\begin{minipage}[t]{0.078\linewidth}
\includegraphics[width=0.68in]{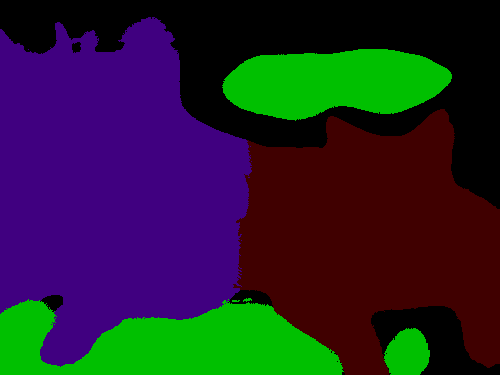}
\end{minipage}
}
\subfigure{
\begin{minipage}[t]{0.078\linewidth}
\includegraphics[width=0.68in]{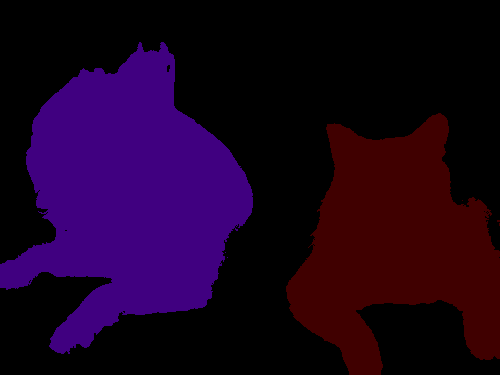}
\end{minipage}
}
\subfigure{
\begin{minipage}[t]{0.078\linewidth}
\includegraphics[width=0.68in]{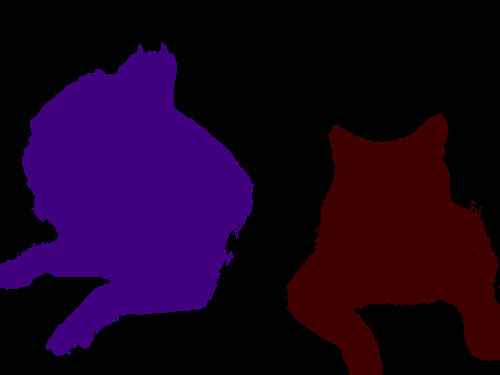}
\end{minipage}
}
\subfigure{
\begin{minipage}[t]{0.078\linewidth}
\includegraphics[width=0.68in]{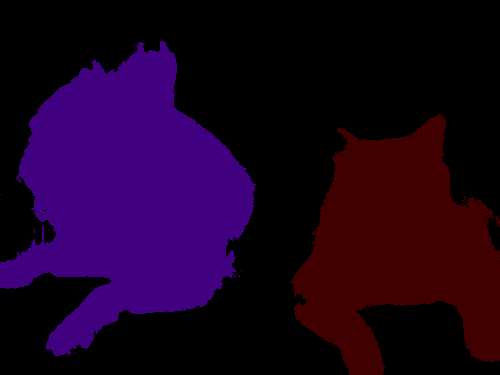}
\end{minipage}
}
\subfigure{
\begin{minipage}[t]{0.078\linewidth}
\includegraphics[width=0.68in]{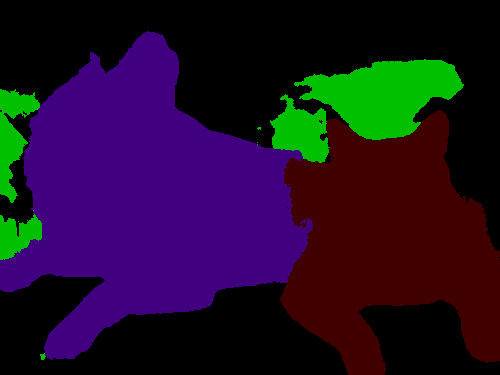}
\end{minipage}
}
\vspace{-8pt}

\subfigure{
\begin{minipage}[t]{0.078\linewidth}
\includegraphics[width=0.68in]{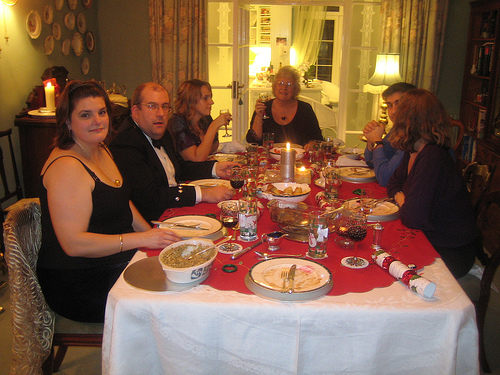}
\end{minipage}
}
\subfigure{
\begin{minipage}[t]{0.078\linewidth}
\includegraphics[width=0.68in]{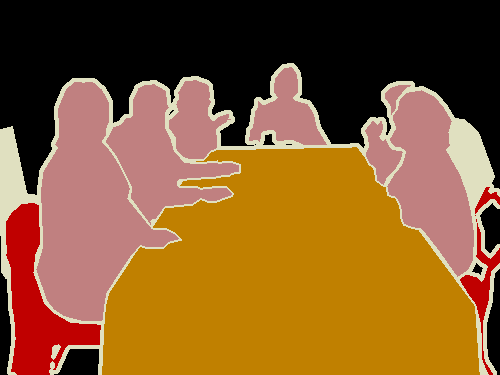}
\end{minipage}
}
\subfigure{
\begin{minipage}[t]{0.078\linewidth}
\includegraphics[width=0.68in]{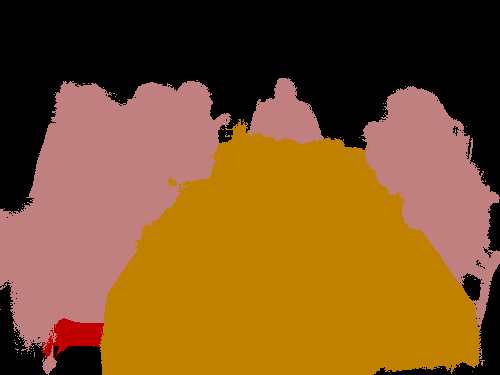}
\end{minipage}
}
\subfigure{
\begin{minipage}[t]{0.078\linewidth}
\includegraphics[width=0.68in]{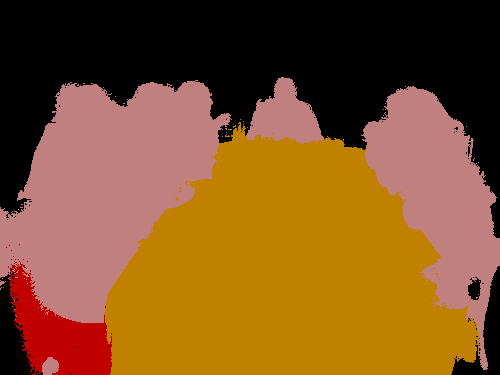}
\end{minipage}
}
\subfigure{
\begin{minipage}[t]{0.078\linewidth}
\includegraphics[width=0.68in]{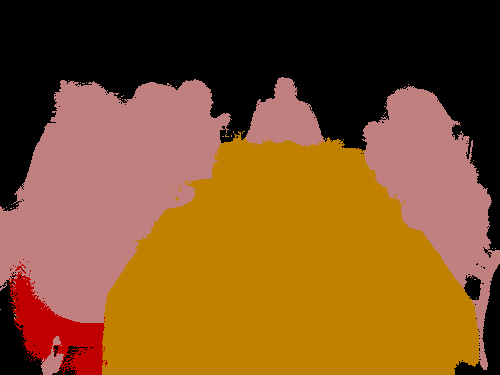}
\end{minipage}
}
\subfigure{
\begin{minipage}[t]{0.078\linewidth}
\includegraphics[width=0.68in]{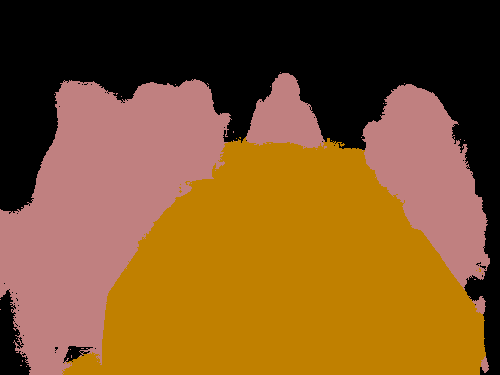}
\end{minipage}
}
\subfigure{
\begin{minipage}[t]{0.078\linewidth}
\includegraphics[width=0.68in]{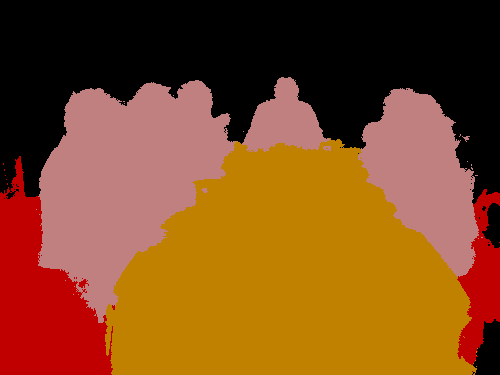}
\end{minipage}
}
\subfigure{
\begin{minipage}[t]{0.078\linewidth}
\includegraphics[width=0.68in]{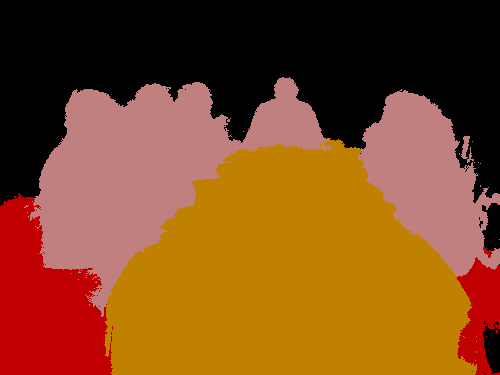}
\end{minipage}
}
\subfigure{
\begin{minipage}[t]{0.078\linewidth}
\includegraphics[width=0.68in]{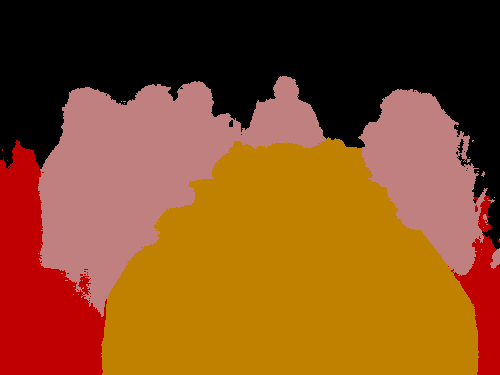}
\end{minipage}
}
\subfigure{
\begin{minipage}[t]{0.078\linewidth}
\includegraphics[width=0.68in]{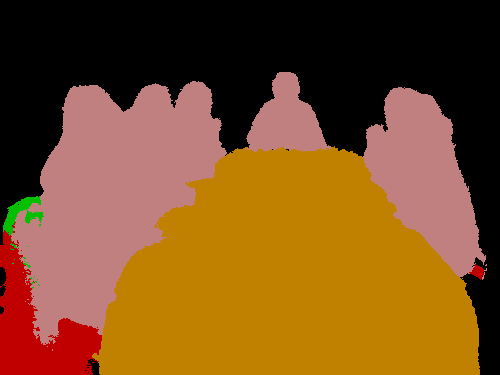}
\end{minipage}
}
\vspace{-8pt}

\subfigure[(a) Image]{
\begin{minipage}[t]{0.078\linewidth}
\includegraphics[width=0.68in]{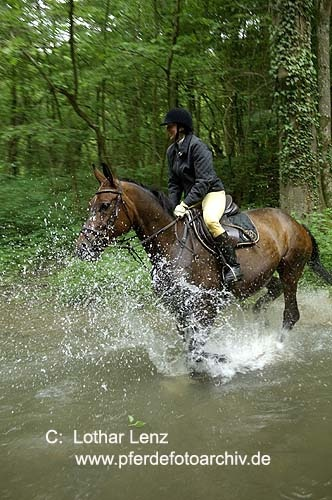}
\end{minipage}
}
\subfigure[(b) GT]{
\begin{minipage}[t]{0.078\linewidth}
\includegraphics[width=0.68in]{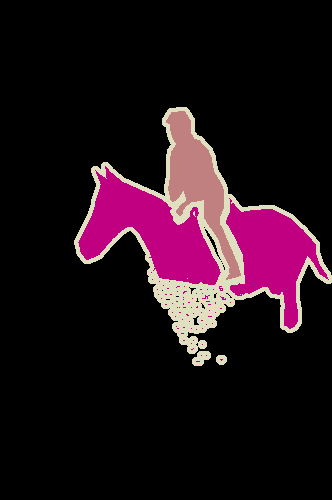}
\end{minipage}
}
\subfigure[(c) V-200]{
\begin{minipage}[t]{0.078\linewidth}
\includegraphics[width=0.68in]{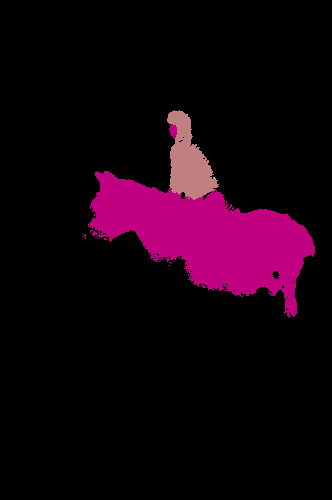}
\end{minipage}
}
\subfigure[(d) V-400]{
\begin{minipage}[t]{0.078\linewidth}
\includegraphics[width=0.68in]{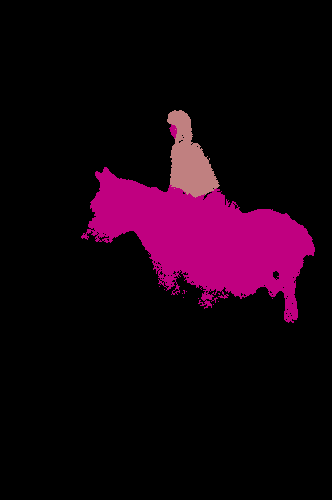}
\end{minipage}
}
\subfigure[(e) V-800]{
\begin{minipage}[t]{0.078\linewidth}
\includegraphics[width=0.68in]{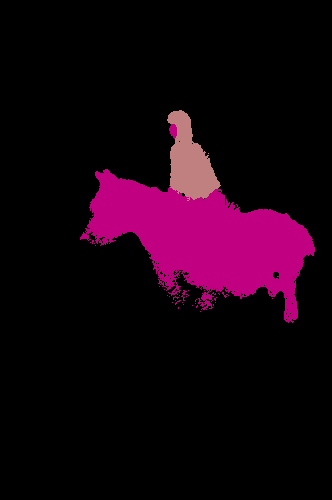}
\end{minipage}
}
\subfigure[(f) V-1464]{
\begin{minipage}[t]{0.078\linewidth}
\includegraphics[width=0.68in]{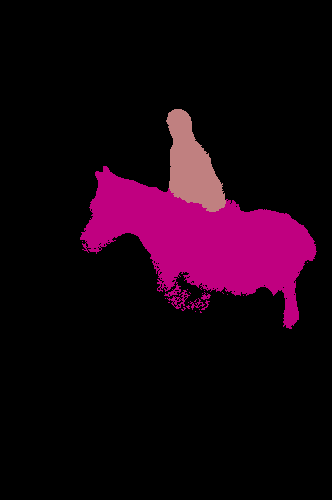}
\end{minipage}
}
\subfigure[(g) R-200]{
\begin{minipage}[t]{0.078\linewidth}
\includegraphics[width=0.68in]{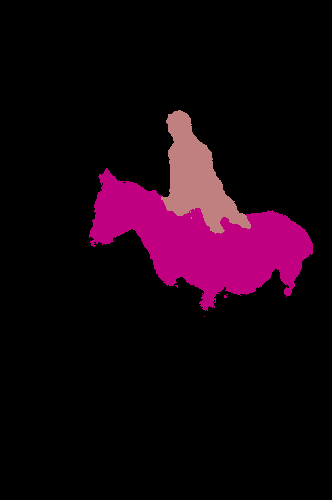}
\end{minipage}
}
\subfigure[(h) R-400]{
\begin{minipage}[t]{0.078\linewidth}
\includegraphics[width=0.68in]{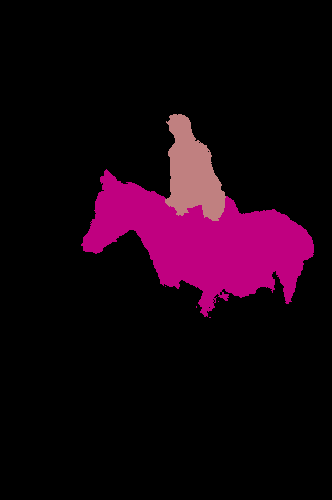}
\end{minipage}
}
\subfigure[(i) R-800]{
\begin{minipage}[t]{0.078\linewidth}
\includegraphics[width=0.68in]{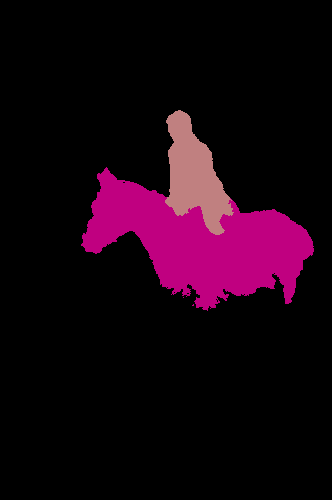}
\end{minipage}
}
\subfigure[(j) R-1464]{
\begin{minipage}[t]{0.078\linewidth}
\includegraphics[width=0.68in]{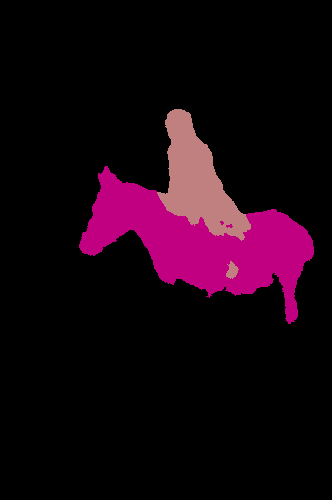}
\end{minipage}
}
\caption{Qualitative results on the PASCAL voc 2012 \emph{val} set using Deeplab v1-vgg16 and Deeplab v2-resnet101 segmentation network. (a) Input images. (b) Ground truth. (c)-(f) are the segmentation results on ${{\mathcal{D}}^{c}}$=200, 400, 800 and 1464 using Deeplab v1-vgg16 network. (g)-(j) are the segmentation results on ${{\mathcal{D}}^{c}}$=200, 400, 800 and 1464 using Deeplab v2-resnet101 network. }

\label{voc12_v1_vis}
\end{figure*}

\begin{table}[htbp]
\caption{\centering Comparisons on PASCAL VOC 2012 \emph{val} and test sets.}
\centering
\begin{tabular}{lccl}
\hline
\textbf{Methods}    & \textbf{Backbone}  & \textbf{\emph{val}} & \textbf{test}       \\ \hline
\multicolumn{4}{l}{Supervision: 10.6 k   pixel-level}                 \\
Deeplab             & v1-vgg16            & 67.6       & 68.6       \\
Deeplab             & v2-resnet101        & 76.5       & 79.7       \\ \hline
\multicolumn{4}{l}{Supervision: 1.4k   pixel-level + 9k image-level} \\
GAIN{~\cite{li2018tell}}  & v1-vgg16            & 60.5       & 62.1       \\
DSRG{~\cite{huang2018weakly}} & v1-vgg16            & 64.3       & -          \\
WSSL{~\cite{papandreou1502weakly}}   & v1-vgg16            & 64.6       & 66.2       \\
MDC{~\cite{wei2018revisiting}}  & v1-vgg16            & 65.7       & 67.6       \\
FickleNet{~\cite{lee2019ficklenet}}  & v1-vgg16            & 65.8       & -          \\
Ouali \emph{et al.}~\cite{Ouali_2020_CVPR}       & resnet50            & 73.2       & -          \\
Luo \emph{et al.}~\cite{10.1007/978-3-030-58558-7_46}           & v2-resnet101        & 76.6       & 77.1       \\ \hline
Ours                & v1-vgg16            & 67.5       & 67.9       \\
Ours                & v2-resnet101        & \textbf{77.1}       & \textbf{77.2}      \\ \hline
\label{Comparison}
\end{tabular}
\end{table}

\begin{table}[]
\caption{\centering Segmentation results on PASCAL VOC 2012 $val$ set with different sizes of ${{\mathcal{D}}^{c}}$.}
\centering
\begin{tabular}{c|cccc|c}
\toprule
{${{\mathcal{D}}^{c}}$}   & {200}  & {400}  & {800}  & {1464} & {Fully-supervised}             \\ \midrule
{v1-vgg16}  & {46.8} & {52.9} & {56.6} & {62.5} & {}                       \\ 
{Ours}  & {63.7} & {64.0} & {64.9} & {67.5} &
 \multirow{-2}{*}{{67.6}} \\ \bottomrule
 {v2-resnet101}  & {58.3} & {63.0} & {68.1} & {74.6} & {}                       \\ 
{Ours}  & {69.5} & {69.7} & {71.3} & {77.1} &
 \multirow{-2}{*}{{76.5}} \\ \bottomrule
\end{tabular}
\label{different size v1v2}
\end{table}

Subsequently, we evaluate our approach with different sizes of fully-labeled set ${{\mathcal{D}}^{c}}$. Similar to~\cite{msibrahiCVPR20segmentcorrectionnet}, we use 200, 400, 800 and 1464 images with strong annotations to train the clean model $\mathcal{C}$ respectively, and compare to the different baselines which are trained only on ${{\mathcal{D}}^{c}}$. As shown in Table~\ref{different size v1v2}, our method achieves an improvement of 5.0\% to 16.9\% (v1-vgg16) and 2.5\% to 11.2\% (v2-resnet101) over the baselines for different sizes splits. More surprisingly, our performance on \emph{val} set outperforms the fully-supervised model (76.5\%) by 0.6\% using v2-resnet101 segmentation network. Notably, the approach works well even with only 1.89\% (${{\mathcal{D}}^{c}}$ = 200) of fully-labeled images, which illustrates that our method can provide a good training signal for the weak set so as to largely alleviate the problem of lacking of segmentation labels. The quantitative results on per-class IoU are shown in Table~\ref{per-class voc v1} and~\ref{per-class voc v2}, and the qualitative results are shown in Figure~\ref{voc12_v1_vis}.

\textbf{Ablation study.} We explore the performance of the generated pseudo labels on 1.4K/9K split in the stages of: 1) Generating pixel-level labels from weak annotations; 2) Detecting clean and noisy labels; 3) Correcting noisy labels. The performance of the generated pseudo labels are measured by the segmentation results using the Deeplab v1-vgg16 model. We train the DeepLab model without and with 1.4K strongly annotated images respectively, and evaluate the mIoU on the validation set.

As shown in Table \ref{Ablation study}, the segmentation result of initial CAM is 57.4\%. For the first strategy (w/ detection and w/o correction), the 1.4K strongly annotated images are applied to train a clean model $\mathcal{C}$, then we use $\mathcal{C}$ to detect the clean labels from initial CAM and train a segmentation network with these clean labels, the performance is boosted by 5.4\% (57.4\% vs. 62.8\%). The result demonstrates that our detecting method can distinguish clean labels from noisy labels. The result also verifies the effect of noisy labels: 
deep models can learn effective information from the limited clean labels, and however the presence of noisy labels dramatically harms the generalization of deep models.

For the second strategy (w/o detection and w/ correction), we just select clean labels according to the activation scores from initial CAM. Similar to~\cite{ahn2019weakly}, we consider pixels with activation scores larger than 0.3 and smaller than 0.05 as foreground and background clean labels, and use them as the supervision for training GAT. Such strategy is similar with~\cite{pu2018graphnet}, but using GAT instead of GCN. However, the segmentation result is only 59.6\%, which is much lower than that of using our proposed detection strategy. The result illustrates that the seed
regions with high confidence contain many mislabeled pixels and~\cite{pu2018graphnet} is not suitable for image-level supervisions. For the third strategy (w/ detection and w/ correction), our approach achieves the best segmentation result of 67.5\%. This clearly validates the effectiveness of the proposed detection and correction method.

\begin{table}[]
\caption{\centering Performance of the pseudo labels during different stages.}
\centering
\begin{tabular}{@{}ccccl@{}}
\toprule
Methods     & w/ detection              & w/ correction             & 9K     & 10.6K\\ \midrule
initial CAM & -                         & -                         & 49.7   & 57.4\\ \midrule
            & { \checkmark} & -                         & 59.8    & 62.8\\
            & -                         & { \checkmark} & 56.9   & 59.6\\
            & { \checkmark} & { \checkmark} & 65.6  & 67.5\\ \midrule
Oracle      & { \checkmark} & { \checkmark} & 66.3  & 68.2\\ \bottomrule
\end{tabular}
\label{Ablation study}
\end{table}

\begin{figure}[]
\centering
\subfigure{
\begin{minipage}[t]{0.296\linewidth}
\centering
\includegraphics[width=1in]{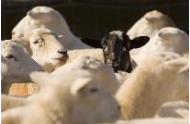}
\end{minipage}
}
\subfigure{
\begin{minipage}[t]{0.296\linewidth}
\centering
\includegraphics[width=1in]{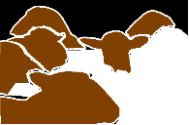}
\end{minipage}
}
\subfigure{
\begin{minipage}[t]{0.296\linewidth}
\centering
\includegraphics[width=1in]{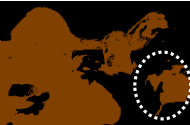}
\end{minipage}
}

\subfigure{
\begin{minipage}[t]{0.296\linewidth}
\centering
\includegraphics[width=1in]{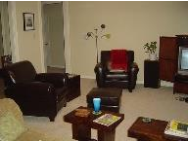}
\end{minipage}
}
\subfigure{
\begin{minipage}[t]{0.296\linewidth}
\centering
\includegraphics[width=1in]{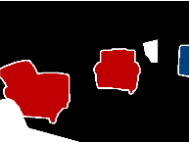}
\end{minipage}
}
\subfigure{
\begin{minipage}[t]{0.296\linewidth}
\centering
\includegraphics[width=1in]{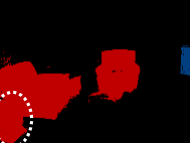}
\end{minipage}
}

\subfigure{
\begin{minipage}[t]{0.296\linewidth}
\centering
\includegraphics[width=1in]{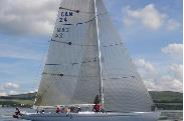}
\end{minipage}
}
\subfigure{
\begin{minipage}[t]{0.296\linewidth}
\centering
\includegraphics[width=1in]{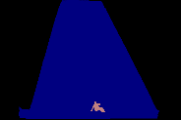}
\end{minipage}
}
\subfigure{
\begin{minipage}[t]{0.296\linewidth}
\centering
\includegraphics[width=1in]{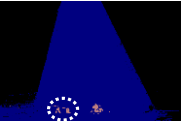}
\end{minipage}
}

\subfigure[(a) Image]{
\begin{minipage}[t]{0.295\linewidth}
\centering
\includegraphics[width=1in]{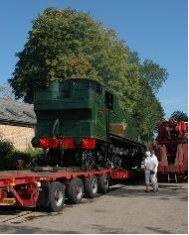}
\end{minipage}
}
\subfigure[(b) Ground truth]{
\begin{minipage}[t]{0.295\linewidth}
\centering
\includegraphics[width=1in]{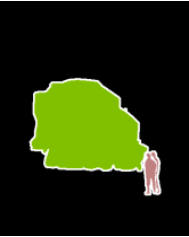}
\end{minipage}
}
\subfigure[(c) Oracle]{
\begin{minipage}[t]{0.295\linewidth}
\centering
\includegraphics[width=1in]{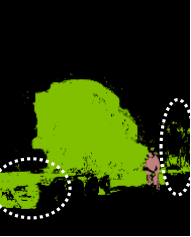}
\end{minipage}
}
\caption{Qualitative results of pseudo labels on the PASCAL VOC 2012 \emph{trainaug} set using oracle strategy, which can successfully correct the label for missing objects in these images (marked by ellipses).}

\label{vis_voc_oracle}
\end{figure}

Besides, it should be interesting to investigate what happens if the selected clean labels are all correct. So we use the ground truth labels to replace the labels of the pixels which are selected by detection strategy and use these ground truth labels to train GAT (hereinafter referred to as the "oracle"). Surprisingly, the segmentation result outperforms the fully-supervised model (68.2\% vs. 67.6\%). This result also verifies  that the key step for seed-and-expansion methods is to generate more accurate seed regions, thus leading to better segmentation performance. Moreover, our proposed graph-based correction method can further correct pixel-level label noise efficiently. Some visual results are given in Figure~\ref{vis_voc_oracle}.

\begin{table}[]
\caption{\centering Performance of the pseudo labels using different number of superpixels and CRF.}
\centering
\begin{tabular}{ccccc}
\toprule
\multicolumn{1}{c}{\multirow{2}{*}{Superpixel}} & \multicolumn{2}{c}{Pseudo label} & \multicolumn{2}{c}{Segmentation} \\  
\multicolumn{1}{c}{}                            & w/o CRF         & w/ CRF         & w/o CRF             & w/ CRF            \\ \midrule
500                                             &{66.1}&{67.5}&{67.1}&{67.2}                   \\
1000                                            &{67.2}&{68.3}&{67.5}&{67.5}                   \\
5000                                            &{68.7}&{69.2}&{67.6}&{67.1}                   \\ \bottomrule
\end{tabular}
\label{superpixel}
\end{table}

\textbf{Impact of superpixels and CRF.} To investigate the impact of the number of superpixels and CRF, we conduct the experiments by over-segmenting one image into 500, 1000 and 5000 superpixels with and without CRF respectively. We observe the quality of pseudo labels in mIoU on the \emph{trainaug} set and the segmentation results on the \emph{val} set using Deeplab v1-vgg16 network. The results are shown in Table~\ref{superpixel}. From the results we can see that the mIoU of pseudo labels increases with the increased number of superpixels, and using dense CRF can further refine them. Meanwhile, the segmentation result is optimal when the images are over-segment into 5000 superpixels and the pseudo labels without using CRF, which is equal to the performance of fully-supervised model. Considering the efficiency and performance, one image is over-segment into about 1000 superpixels in our experiments.

\begin{figure}[]
\centering
\subfigure[${{\mathcal{D}}^{c}}$=200]{
\begin{minipage}[t]{0.45\linewidth}
\centering
\includegraphics[width=1.7in]{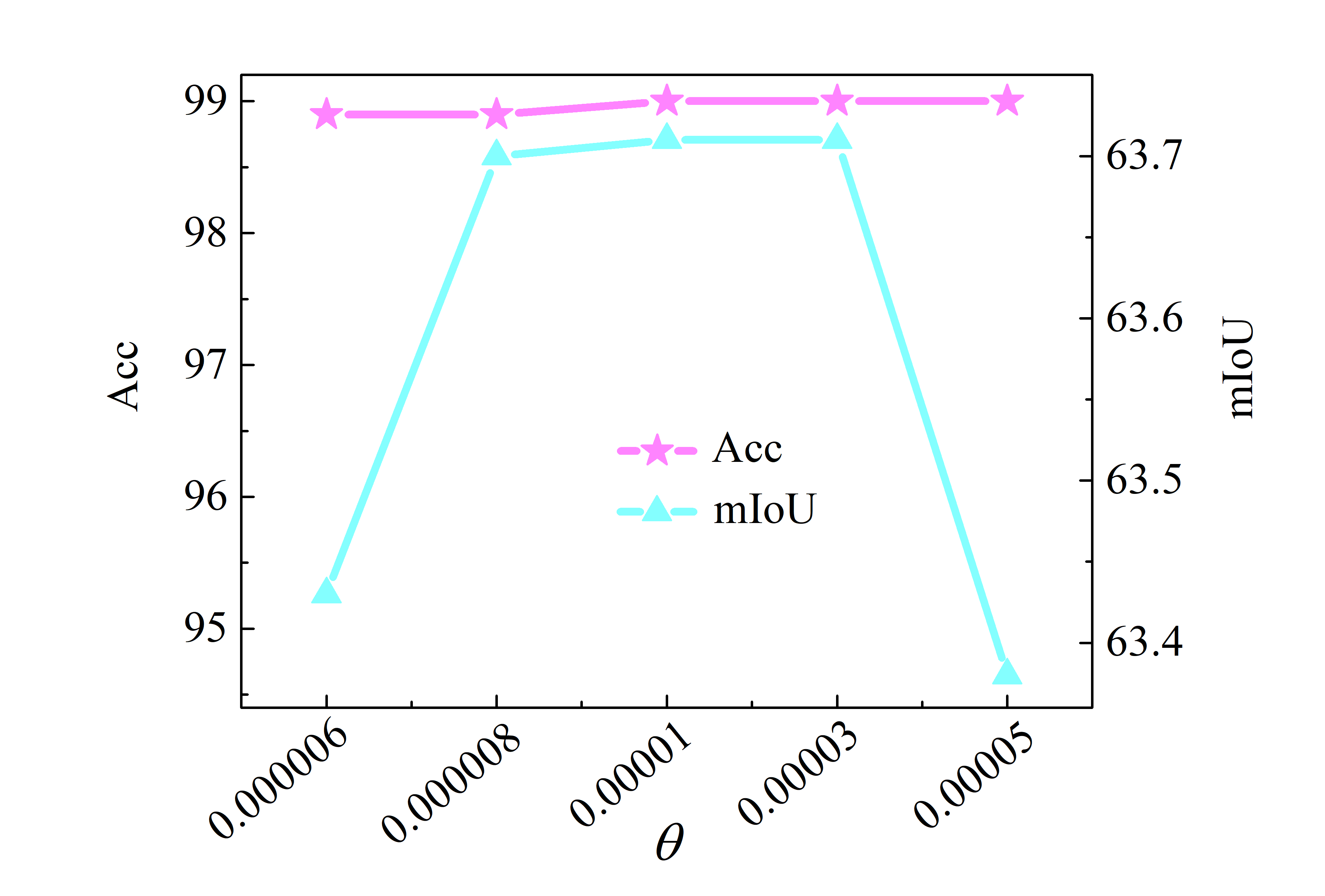}
\end{minipage}
}
\subfigure[${{\mathcal{D}}^{c}}$=400]{
\begin{minipage}[t]{0.45\linewidth}
\centering
\includegraphics[width=1.7in]{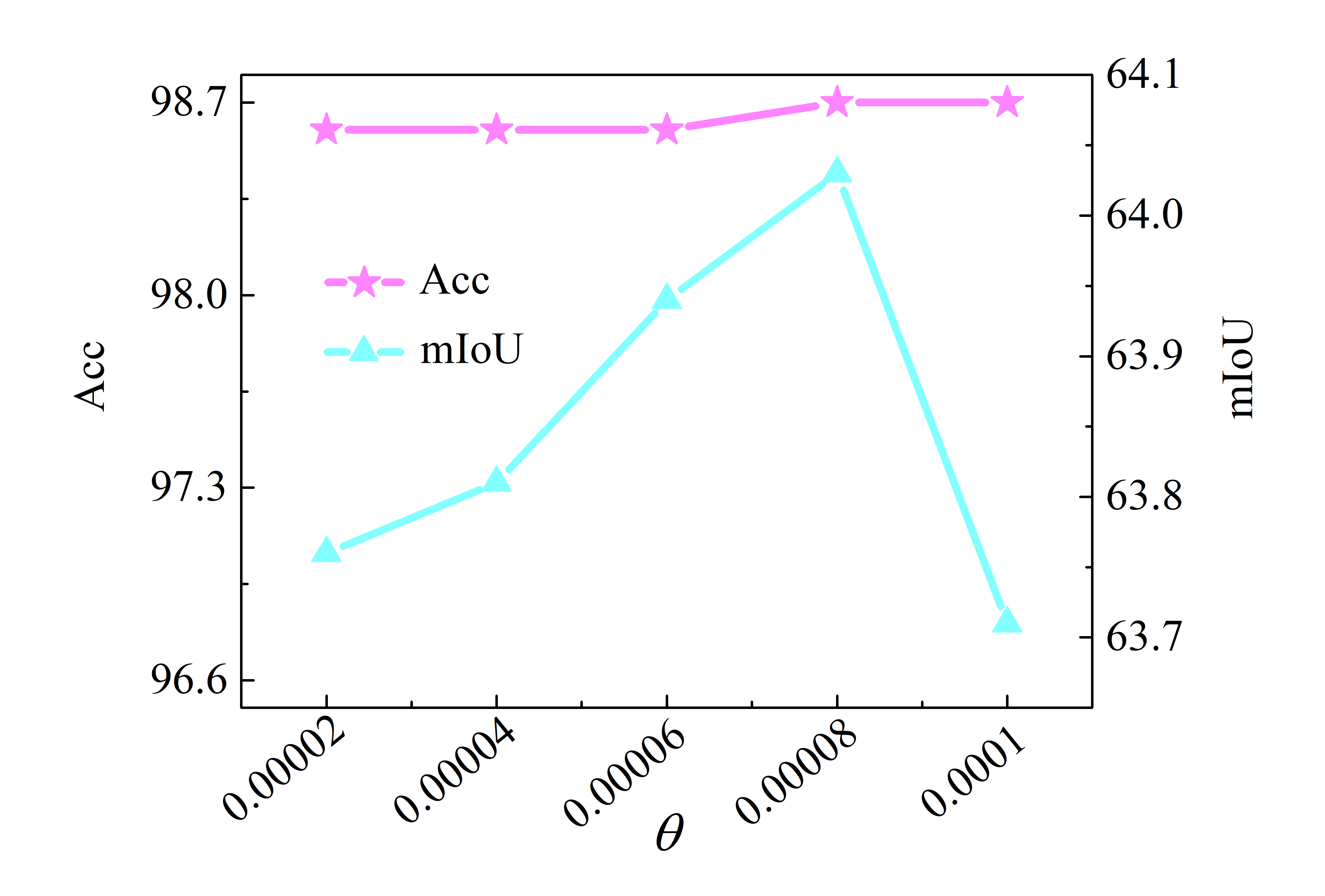}
\end{minipage}
}

\subfigure[${{\mathcal{D}}^{c}}$=800]{
\begin{minipage}[t]{0.45\linewidth}
\centering
\includegraphics[width=1.7in]{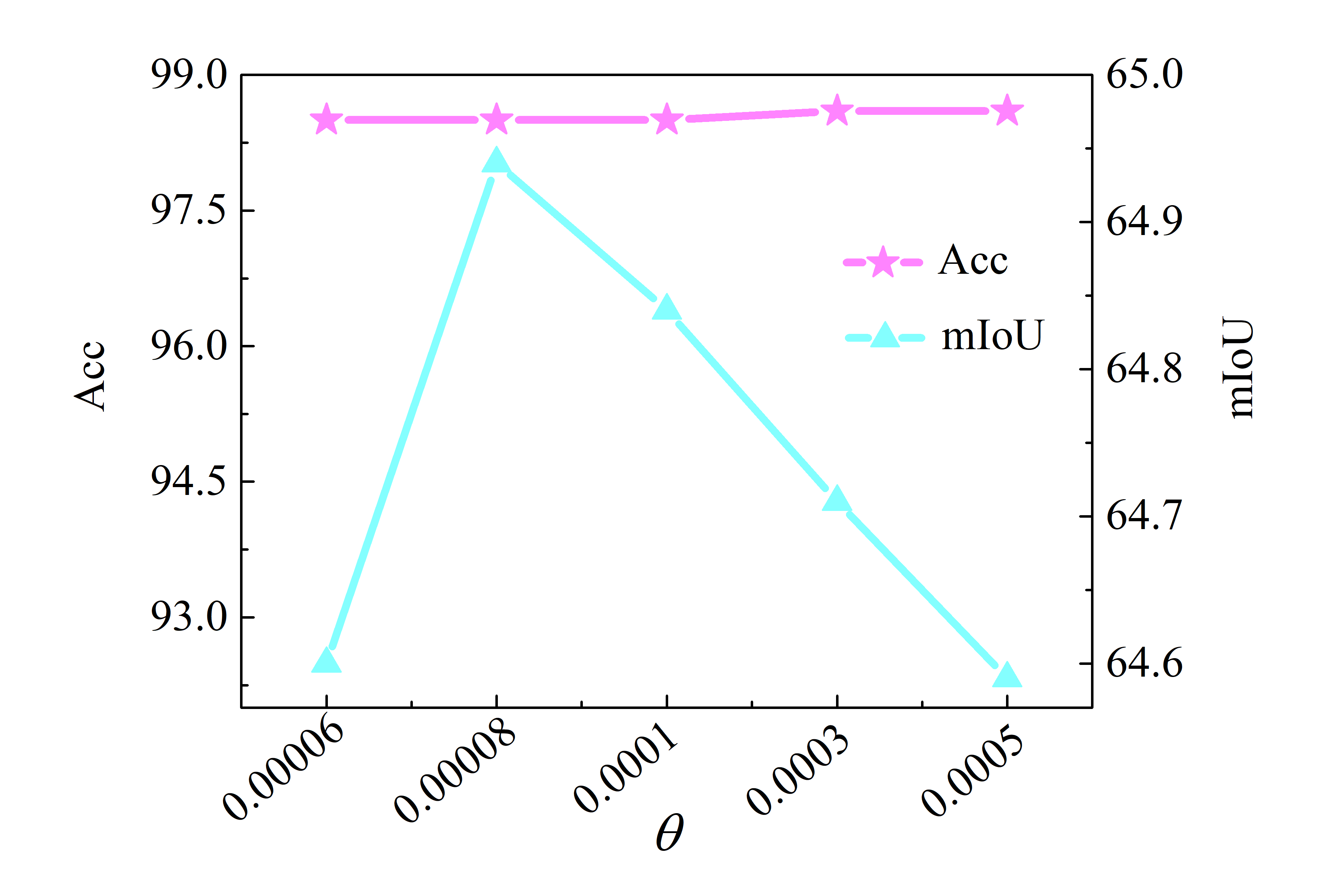}
\end{minipage}
}
\subfigure[${{\mathcal{D}}^{c}}$=1464]{
\begin{minipage}[t]{0.45\linewidth}
\centering
\includegraphics[width=1.7in]{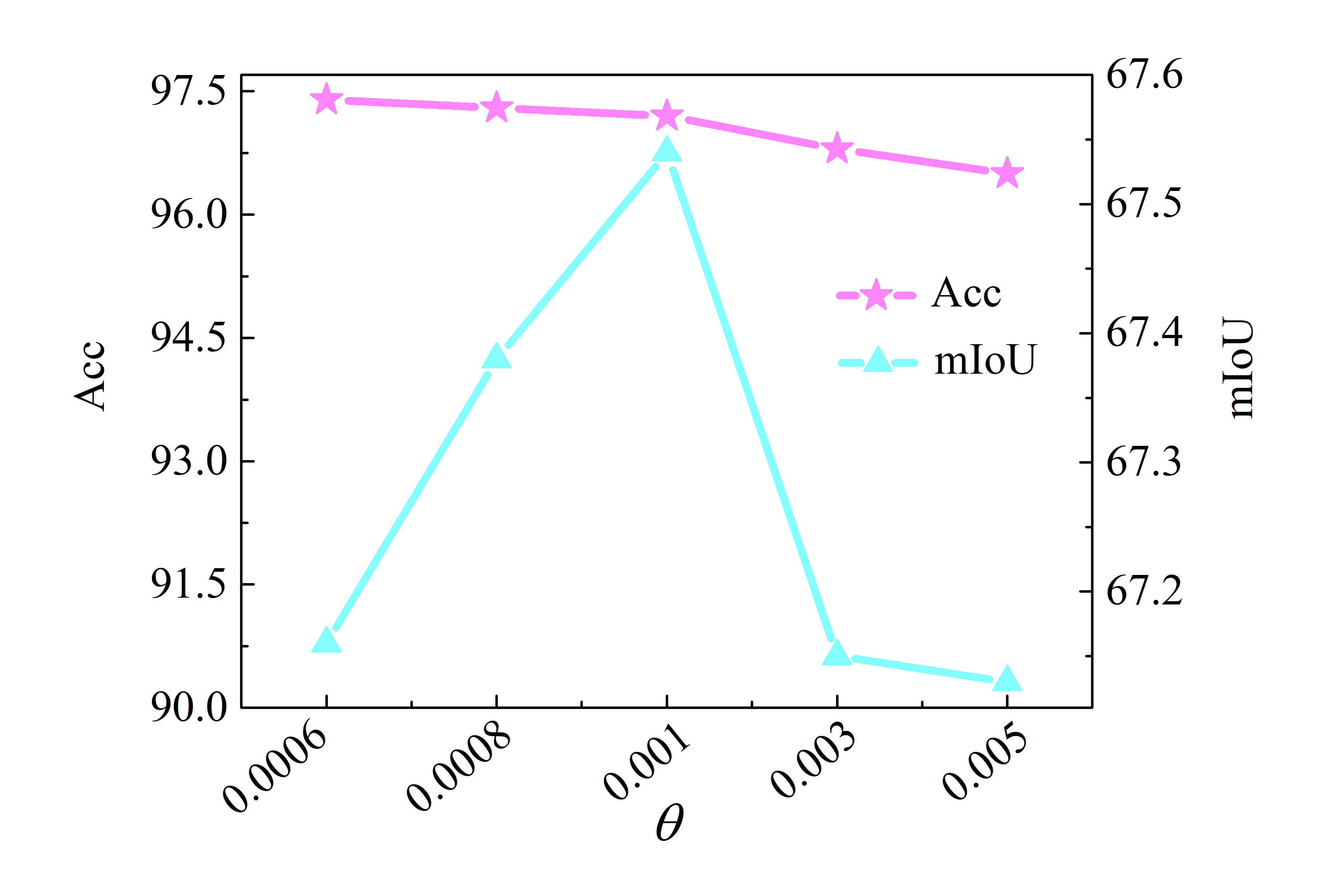}
\end{minipage}
}
\caption{The accuracy of the selected clean labels and the mIoU of the segmentation results under different $\theta$.}
\label{threshold}
\end{figure}

\begin{figure}[]
\centering
\includegraphics[width=0.9\linewidth]{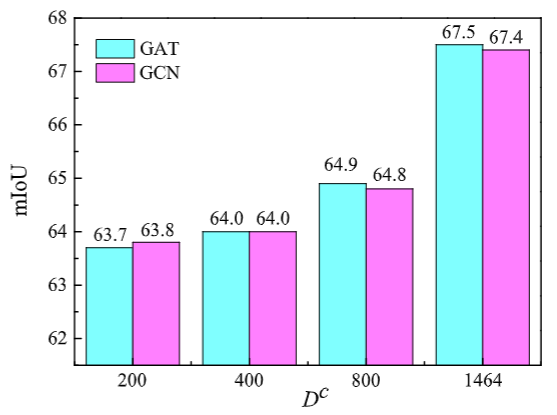}
\caption{The comparison of segmentation results using GCN or GAT to perform noisy labels correction.}
\label{GCNvsGAT}
\end{figure}

\textbf{Impact of parameter $\theta$.}
The parameter $\theta$ is the key to select clean labels from initial pixel-level noisy labels, so we investigate the impact of $\theta$ on different splits using Deeplab v1-vgg16 segmentation network. Since we have the ground truth labels on ${{\mathcal{D}}^{c}}$, the accuracy of the selected clean labels are measured on ${{\mathcal{D}}^{c}}$, and the segmentation results are also reported on the validation set. As shown in Figure~\ref{threshold}, we select $\theta$ from 0.0006 to 0.005 on strong/weak split of 1.4K/9K dataset, the accuracy of the selected clean labels evaluated on 1.4K subset drops from 97.4\% to 96.5\%, and the segmentation result first rises from 67.16\% to 67.54\%, and then drops to 67.13\%. When $\theta$ = 0.001, the pixel accuracy achieves 97.2\% and the segmentation result also reaches 67.54\%. Therefore, $\theta$ is set to 0.001 on 1.4k/9k split. Note that we can use this strategy to select a suitable threshold on different settings. In our experiments, $\theta$ takes 0.00003, 0.00008 and 0.00008  under the setting of ${{\mathcal{D}}^{c}}$=200, 400 and 800 respectively.

\textbf{Correcting noisy labels by GAT vs. GCN.} To investigate the performance of correcting noisy labels using different graph neural networks, we apply GCN instead of GAT to perform noisy label correction. Figure \ref{GCNvsGAT} shows the comparison results. The segmentation results which using GCN are 63.8\%, 64.0\% , 64.8\% and 67.4\% under the setting of ${{\mathcal{D}}^{c}}$=200, 400, 800 and 1464. It also achieves a performance improvement of 4.9\% to 17\% over the baseline for different sizes splits. Overall, GAT slightly achieves better performance than that of GCN in most settings. 

\subsection{Results on PASCAL-Context dataset}
\begin{figure}[]
\subfigure{
\begin{minipage}[t]{0.2\linewidth}
\includegraphics[width=0.8in]{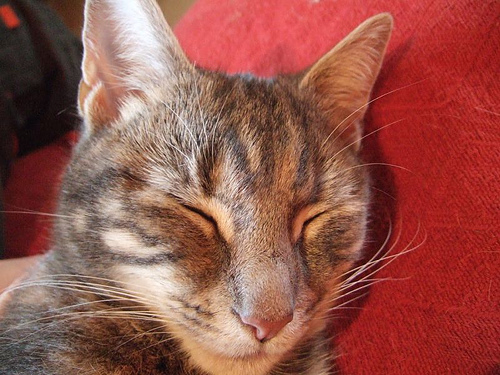}
\end{minipage}
}
\subfigure{
\begin{minipage}[t]{0.2\linewidth}
\includegraphics[width=0.8in]{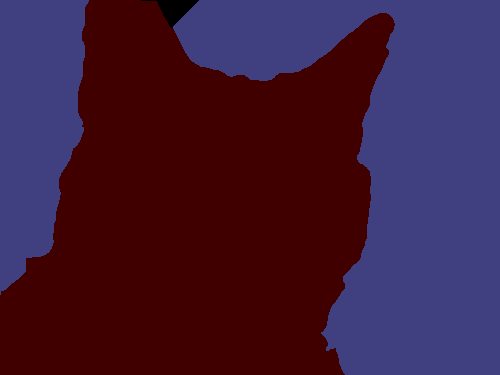}
\end{minipage}
}
\subfigure{
\begin{minipage}[t]{0.2\linewidth}
\includegraphics[width=0.8in]{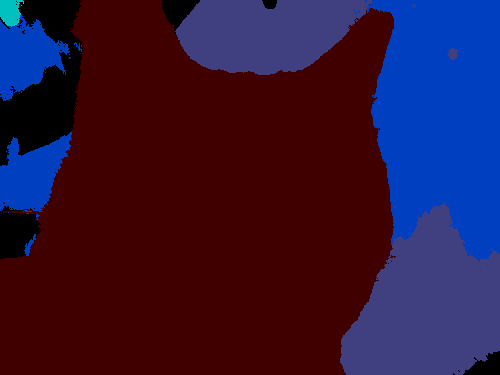}
\end{minipage}
}
\subfigure{
\begin{minipage}[t]{0.2\linewidth}
\includegraphics[width=0.8in]{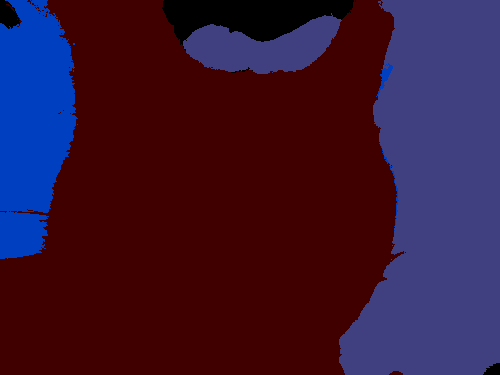}
\end{minipage}
}
\vspace{-7pt}

\subfigure{
\begin{minipage}[t]{0.2\linewidth}
\includegraphics[width=0.8in]{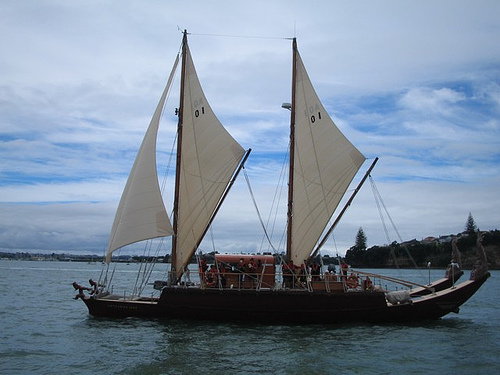}
\end{minipage}
}
\subfigure{
\begin{minipage}[t]{0.2\linewidth}
\includegraphics[width=0.8in]{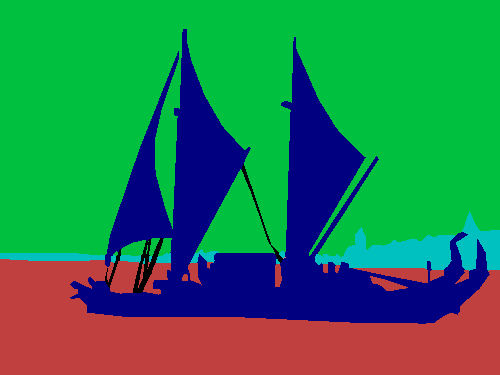}
\end{minipage}
}
\subfigure{
\begin{minipage}[t]{0.2\linewidth}
\includegraphics[width=0.8in]{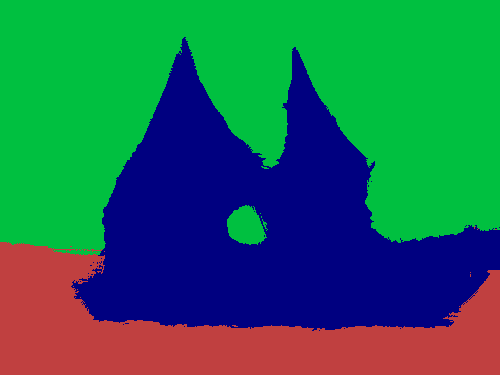}
\end{minipage}
}
\subfigure{
\begin{minipage}[t]{0.2\linewidth}
\includegraphics[width=0.8in]{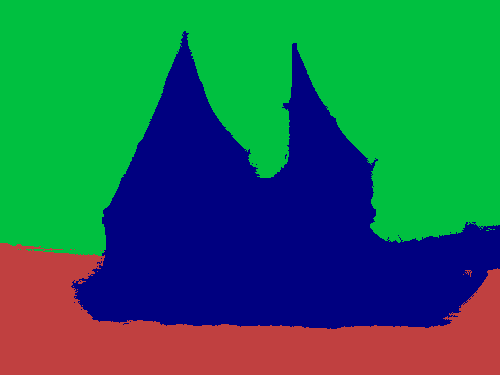}
\end{minipage}
}
\vspace{-7pt}

\subfigure{
\begin{minipage}[t]{0.2\linewidth}
\includegraphics[width=0.8in]{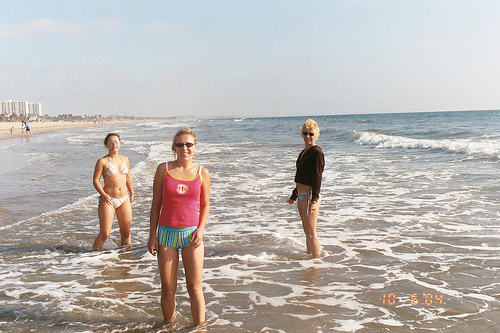}
\end{minipage}
}
\subfigure{
\begin{minipage}[t]{0.2\linewidth}
\includegraphics[width=0.8in]{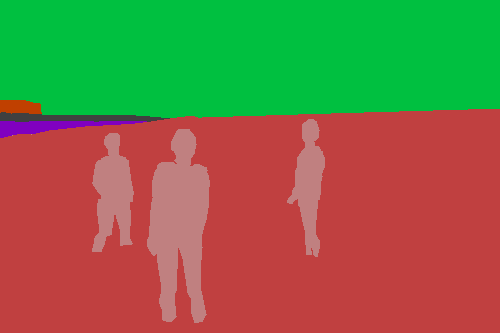}
\end{minipage}
}
\subfigure{
\begin{minipage}[t]{0.2\linewidth}
\includegraphics[width=0.8in]{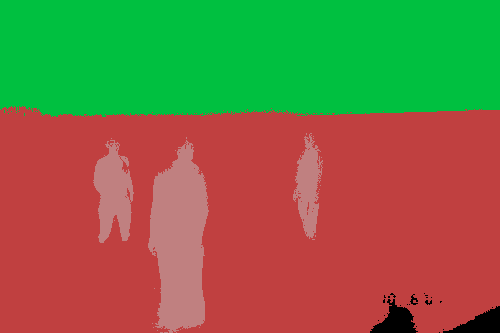}
\end{minipage}
}
\subfigure{
\begin{minipage}[t]{0.2\linewidth}
\includegraphics[width=0.8in]{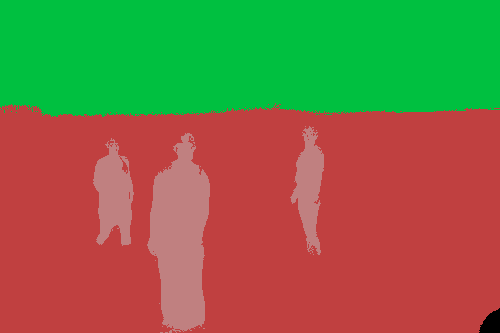}
\end{minipage}
}
\vspace{-7pt}

\subfigure[(a) Image]{
\begin{minipage}[t]{0.2\linewidth}
\includegraphics[width=0.8in]{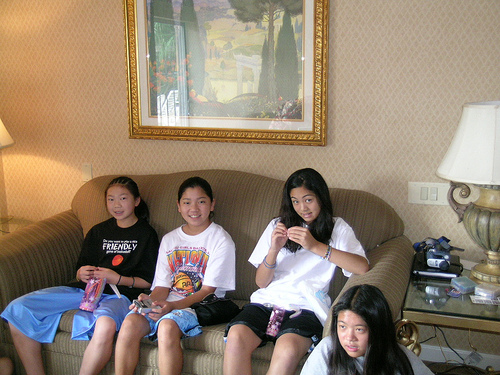}
\end{minipage}
}
\subfigure[(b) Ground truth]{
\begin{minipage}[t]{0.2\linewidth}
\includegraphics[width=0.8in]{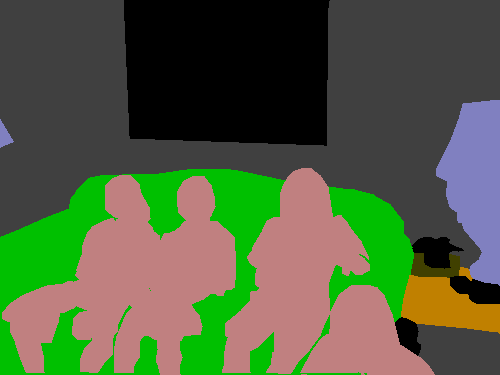}
\end{minipage}
}
\subfigure[(c) ${{D}^{c}}=625$]{
\begin{minipage}[t]{0.2\linewidth}
\includegraphics[width=0.8in]{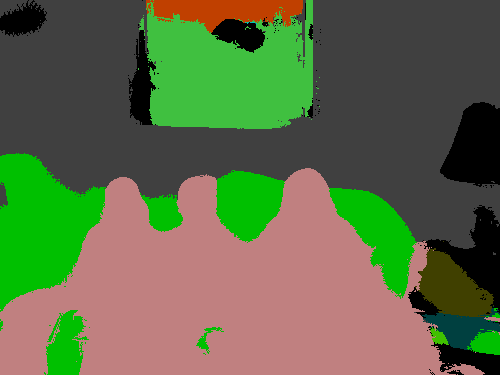}
\end{minipage}
}
\subfigure[(d) ${{D}^{c}}=1250$]{
\begin{minipage}[t]{0.2\linewidth}
\includegraphics[width=0.8in]{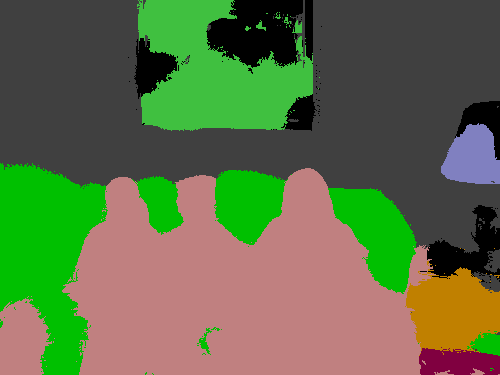}
\end{minipage}
}
\caption{Qualitative results on the PASCAL-Context test set using Deeplab v1-vgg16 segmentation network. (a) Input images. (b) Ground truth. (c) and (d) are the segmentation results on ${{D}^{c}}=$625 and 1250. }

\label{vis_context}
\end{figure}

Our approach successfully generalizes to the whole scene parsing PASCAL-Context dataset using Deeplab v1-vgg16 segmentation network. Table~\ref{ContextResults} shows the performance on two splits (1/8 and 1/4 strongly annotated images) on PASCAL-Context dataset. Although this dataset is smaller and more difficult than PASCAL VOC 2012, there is still an improvement over the baseline of 5.0\% and 2.3\% for the 1/8 and 1/4 splits, respectively. The qualitative results are shown in Figure~\ref{vis_context}.

\begin{table}[]
\centering
\caption{\centering Segmentation results on PASCAL-Context test set with different sizes of ${{\mathcal{D}}^{c}}$ using Deeplab v1-vgg16 segmentation network.}
\centering
\begin{tabular}{@{}c|cc|c@{}}
\toprule
{${{D}_{c}}$}& {625} & {1250} & { Fully-supervised  }             \\ \midrule
{ DeepLab }         & {30.8} & {35.0}                    \\
{ Ours } & {35.8} & {37.3} & \multirow{-2}{*}{{38.6}} \\ \bottomrule
\end{tabular}
\label{ContextResults}
\end{table}
\begin{table}[]
\setlength{\abovecaptionskip}{0cm}
\small
\caption{\centering Segmentation results on MS-COCO $val$ set with different sizes of ${{\mathcal{D}}^{c}}$ using Deeplab v1-vgg16 segmentation network.}
\centering
\begin{tabular}{@{}c|ccc|c@{}}
\toprule
{${{D}_{c}}$}& {1293} &{10348} & {20696} & { Fully-supervised  }             \\ \midrule
{ DeepLab }     &{23.5}    & {29.1} & {29.3}                    \\
{ Ours } &{29.6}& {31.5} & {31.6} & \multirow{-2}{*}{{29.5}} \\ \bottomrule
\end{tabular}
\label{COCOResults}
\end{table}

\begin{figure}[]
\centering
\subfigure{
\begin{minipage}[t]{0.296\linewidth}
\centering
\includegraphics[width=1in]{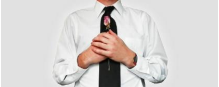}
\end{minipage}
}
\subfigure{
\begin{minipage}[t]{0.296\linewidth}
\centering
\includegraphics[width=1in]{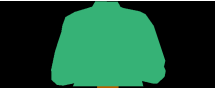}
\end{minipage}
}
\subfigure{
\begin{minipage}[t]{0.296\linewidth}
\centering
\includegraphics[width=1in]{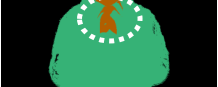}
\end{minipage}
}

\subfigure{
\begin{minipage}[t]{0.296\linewidth}
\centering
\includegraphics[width=1in]{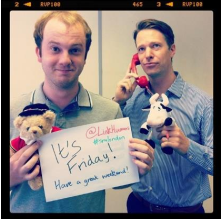}
\end{minipage}
}
\subfigure{
\begin{minipage}[t]{0.296\linewidth}
\centering
\includegraphics[width=1in]{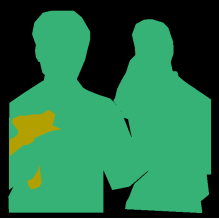}
\end{minipage}
}
\subfigure{
\begin{minipage}[t]{0.296\linewidth}
\centering
\includegraphics[width=1in]{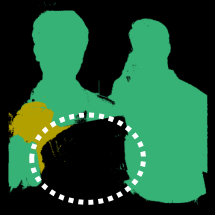}
\end{minipage}
}

\subfigure[(a) Image]{
\begin{minipage}[t]{0.295\linewidth}
\centering
\includegraphics[width=1in]{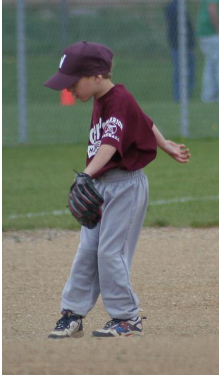}
\end{minipage}
}
\subfigure[(b) Ground truth]{
\begin{minipage}[t]{0.295\linewidth}
\centering
\includegraphics[width=1in]{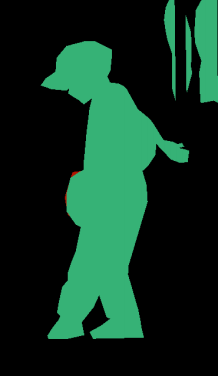}
\end{minipage}
}
\subfigure[(c) Ours]{
\begin{minipage}[t]{0.295\linewidth}
\centering
\includegraphics[width=1in]{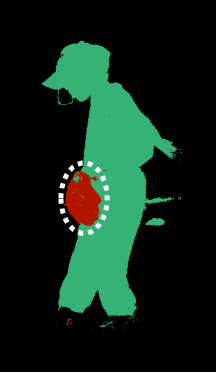}
\end{minipage}
}
\caption{Qualitative results on the MS-COCO training set, our method can successfully correct the labels for missing objects in ground truth images (marked by ellipses).}

\label{vis_coco}
\end{figure}
\vspace{-15pt}

\begin{figure}[h]
\centering
\subfigure{
\begin{minipage}[t]{0.15\linewidth}
\includegraphics[width=0.65in]{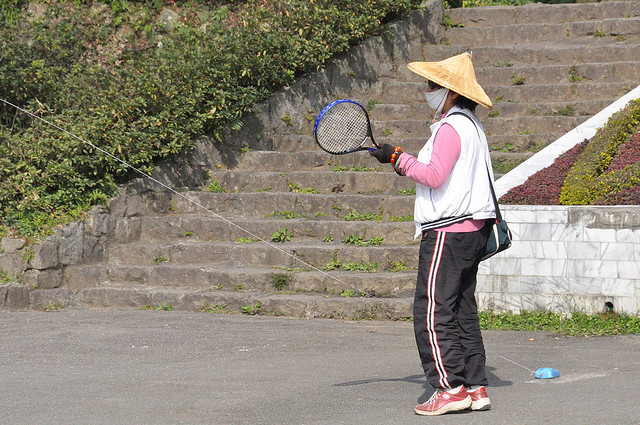}
\end{minipage}
}
\subfigure{
\begin{minipage}[t]{0.15\linewidth}
\includegraphics[width=0.65in]{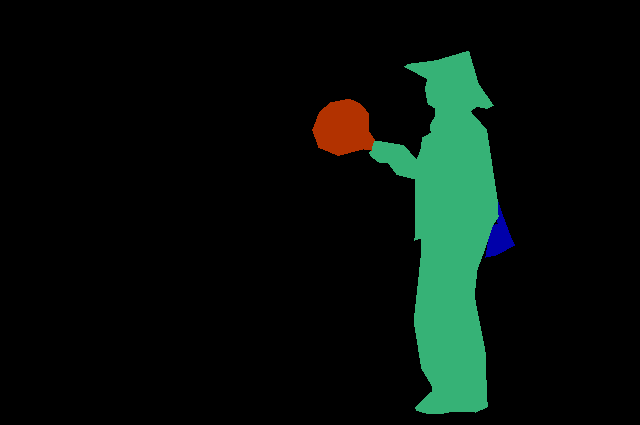}
\end{minipage}
}
\subfigure{
\begin{minipage}[t]{0.15\linewidth}
\includegraphics[width=0.65in]{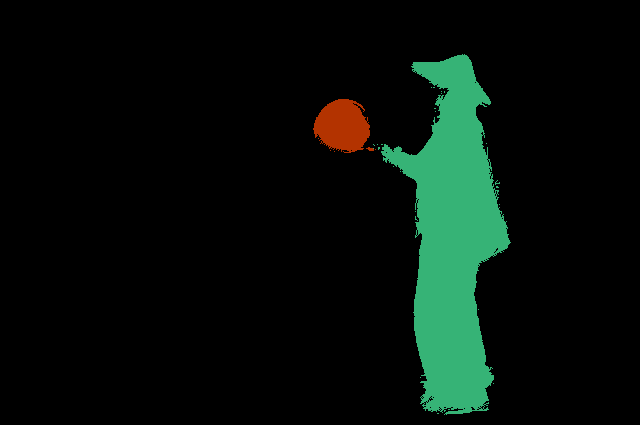}
\end{minipage}
}
\subfigure{
\begin{minipage}[t]{0.15\linewidth}
\includegraphics[width=0.65in]{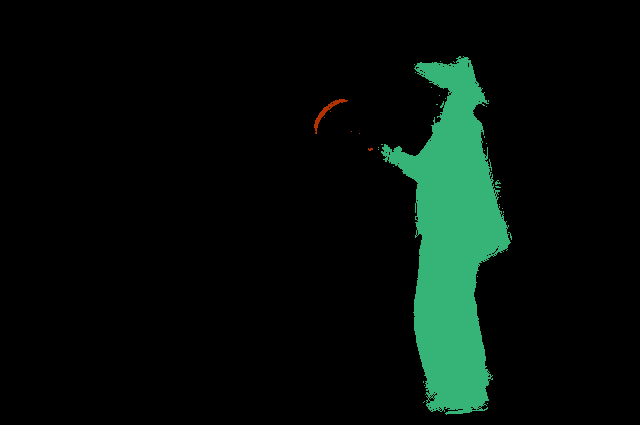}
\end{minipage}
}
\subfigure{
\begin{minipage}[t]{0.15\linewidth}
\includegraphics[width=0.65in]{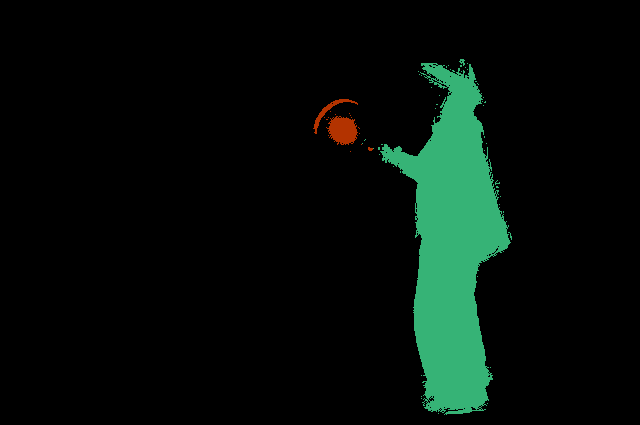}
\end{minipage}
}
\vspace{-8pt}

\subfigure{
\begin{minipage}[t]{0.15\linewidth}
\includegraphics[width=0.65in]{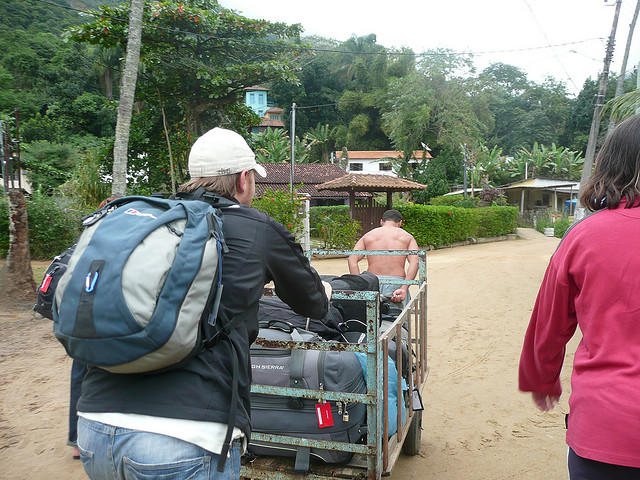}
\end{minipage}
}
\subfigure{
\begin{minipage}[t]{0.15\linewidth}
\includegraphics[width=0.65in]{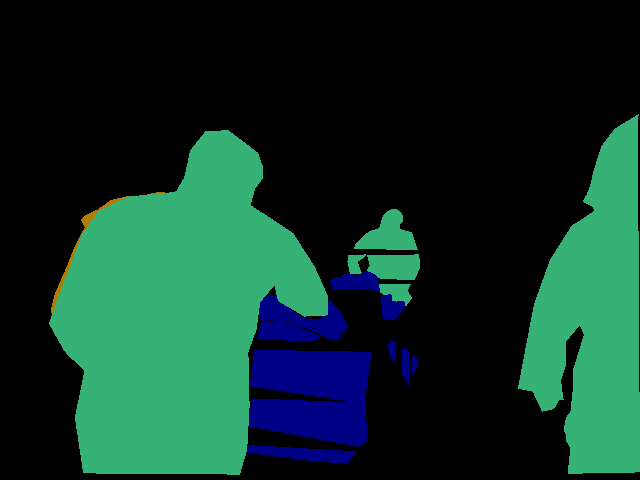}
\end{minipage}
}
\subfigure{
\begin{minipage}[t]{0.15\linewidth}
\includegraphics[width=0.65in]{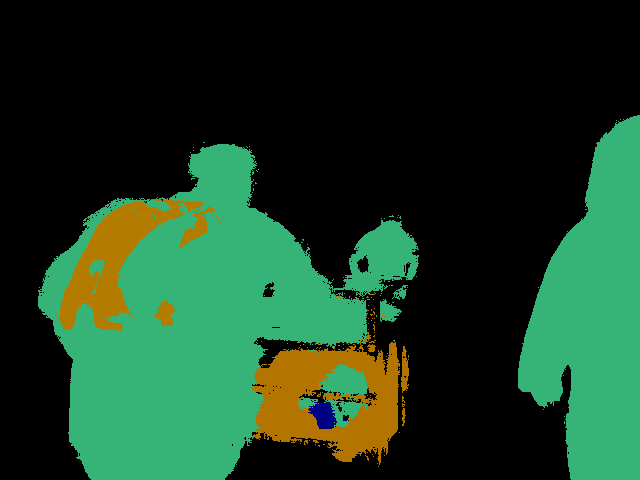}
\end{minipage}
}
\subfigure{
\begin{minipage}[t]{0.15\linewidth}
\includegraphics[width=0.65in]{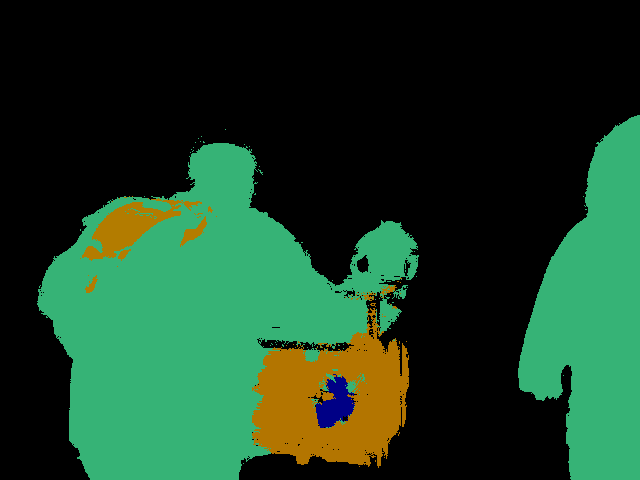}
\end{minipage}
}
\subfigure{
\begin{minipage}[t]{0.15\linewidth}
\includegraphics[width=0.65in]{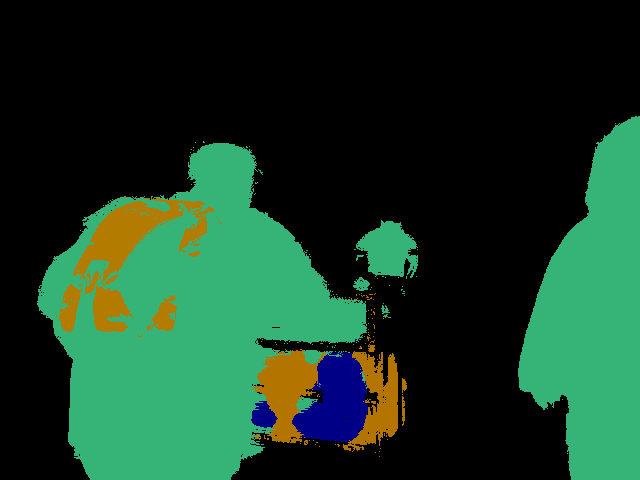}
\end{minipage}
}
\vspace{-8pt}

\subfigure{
\begin{minipage}[t]{0.15\linewidth}
\includegraphics[width=0.65in]{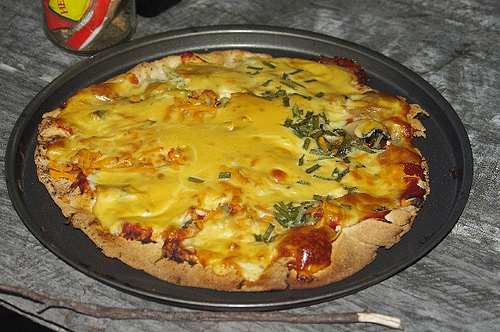}
\end{minipage}
}
\subfigure{
\begin{minipage}[t]{0.15\linewidth}
\includegraphics[width=0.65in]{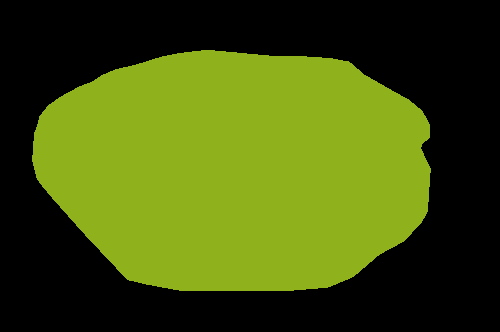}
\end{minipage}
}
\subfigure{
\begin{minipage}[t]{0.15\linewidth}
\includegraphics[width=0.65in]{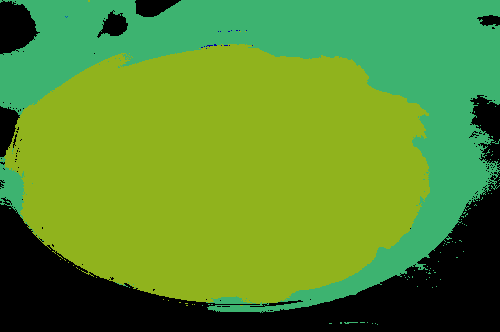}
\end{minipage}
}
\subfigure{
\begin{minipage}[t]{0.15\linewidth}
\includegraphics[width=0.65in]{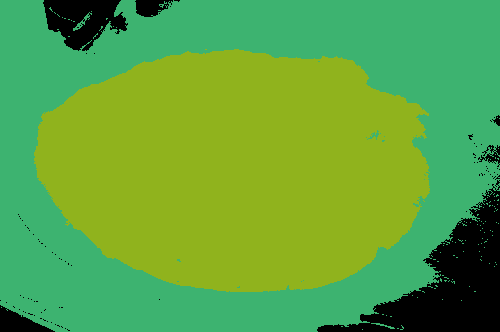}
\end{minipage}
}
\subfigure{
\begin{minipage}[t]{0.15\linewidth}
\includegraphics[width=0.65in]{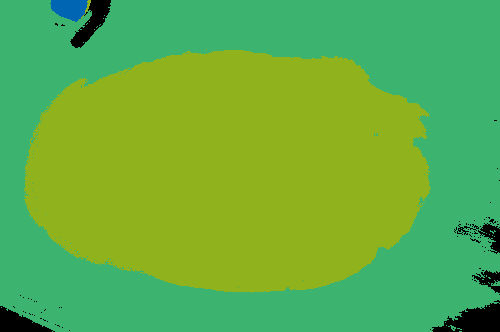}
\end{minipage}
}
\vspace{-8pt}

\subfigure{
\begin{minipage}[t]{0.15\linewidth}
\includegraphics[width=0.65in]{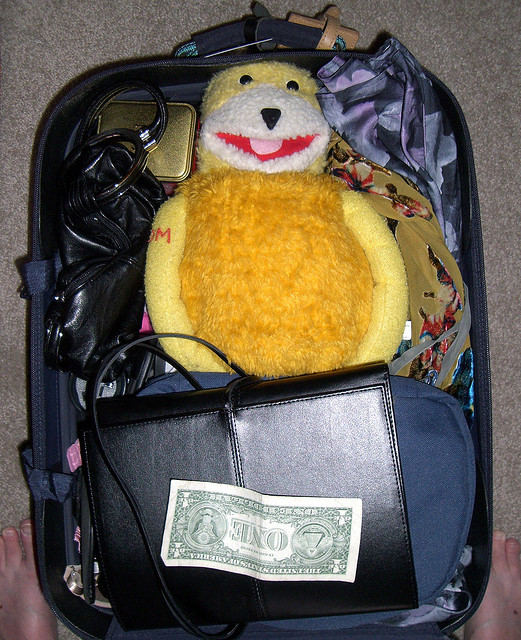}
\end{minipage}
}
\subfigure{
\begin{minipage}[t]{0.15\linewidth}
\includegraphics[width=0.65in]{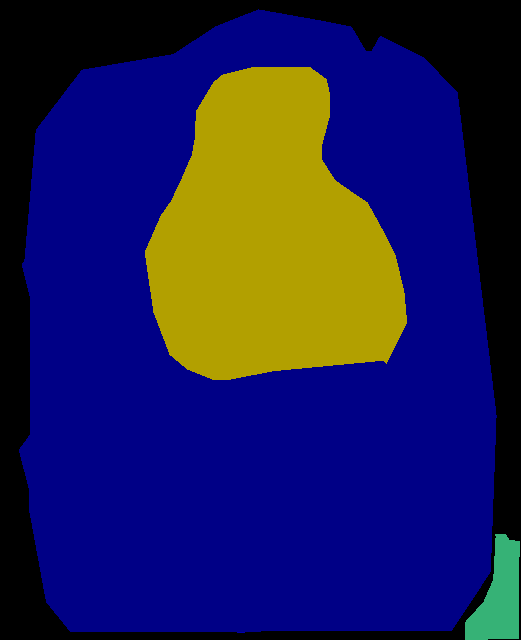}
\end{minipage}
}
\subfigure{
\begin{minipage}[t]{0.15\linewidth}
\includegraphics[width=0.65in]{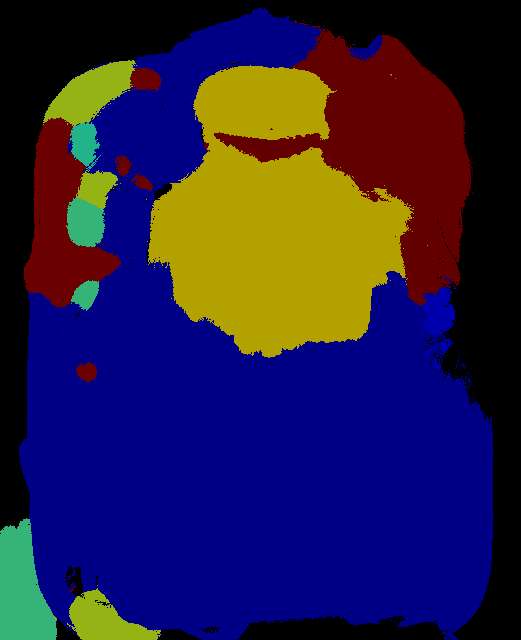}
\end{minipage}
}
\subfigure{
\begin{minipage}[t]{0.15\linewidth}
\includegraphics[width=0.65in]{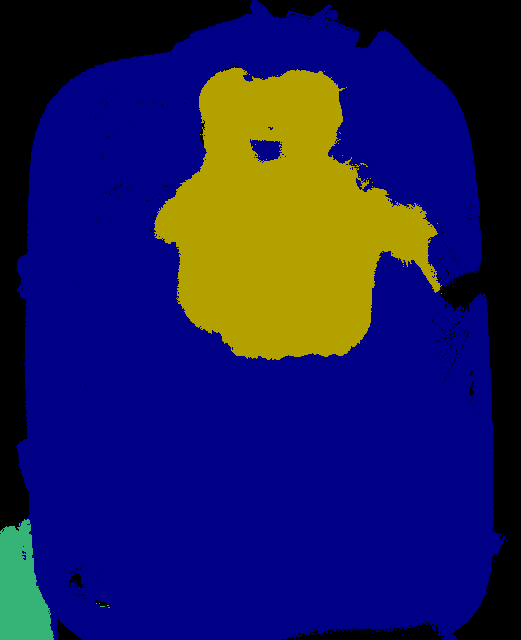}
\end{minipage}
}
\subfigure{
\begin{minipage}[t]{0.15\linewidth}
\includegraphics[width=0.65in]{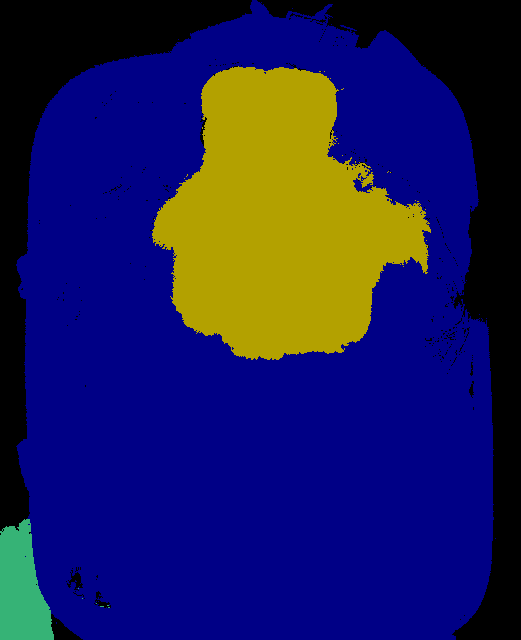}
\end{minipage}
}
\vspace{-8pt}

\subfigure[(a) Image]{
\begin{minipage}[t]{0.15\linewidth}
\includegraphics[width=0.65in]{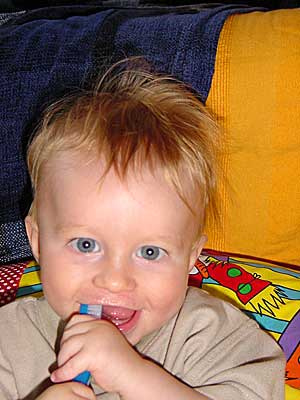}
\end{minipage}
}
\subfigure[(b) GT]{
\begin{minipage}[t]{0.15\linewidth}
\includegraphics[width=0.65in]{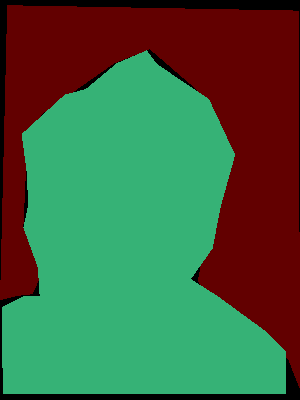}
\end{minipage}
}
\subfigure[(c) 1293]{
\begin{minipage}[t]{0.15\linewidth}
\includegraphics[width=0.65in]{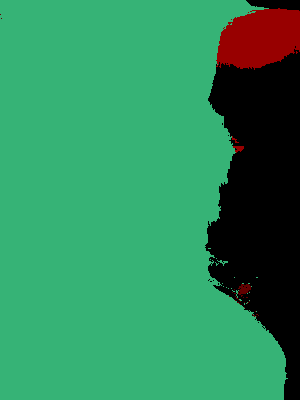}
\end{minipage}
}
\subfigure[(d) 10348]{
\begin{minipage}[t]{0.15\linewidth}
\includegraphics[width=0.65in]{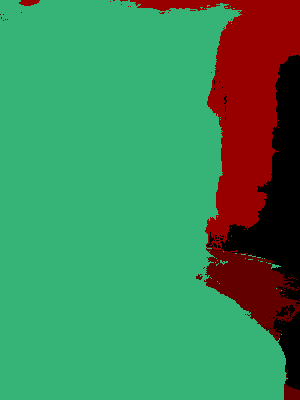}
\end{minipage}
}
\subfigure[(e) 20696]{
\begin{minipage}[t]{0.15\linewidth}
\includegraphics[width=0.65in]{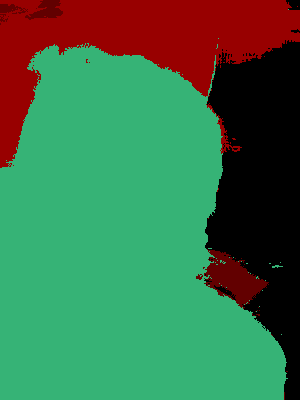}
\end{minipage}
}
\caption{Qualitative results on the MS-COCO \emph{val} set using Deeplab v1-vgg16 segmentation network. (a) Input images. (b) Ground truth. (c)-(e) are the segmentation results on ${{D}^{c}}=$1293, 10348 and 20696.}

\label{vis_coco_segmentation}
\end{figure}
\subsection{Results on MS-COCO dataset}
To further evaluate our proposed method, we conduct experiments on MS-COCO dataset using Deeplab v1-vgg16 segmentation network. Table~\ref{COCOResults} shows the performance on three splits (1/64, 1/8 and 1/4 strongly annotated images). Although this dataset is collected from a complex natural context, there is still an improvement over the baseline of 6.1\%, 2.4\% and 2.3\% for the 1/64, 1/8 and 1/4 splits, respectively. More  surprisingly, our semi-supervised setting outperforms the fully-supervised model only using 1/64 of fully-labeled images. The qualitative results of the pseudo labels generated by our method are shown in Figure~\ref{vis_coco}. Due to the rough labeling of the MS-COCO dataset, the pseudo labels corrected by our method successfully highlight some instances which are improperly labeled in ground truth annotations, thus resulting in better performance. This result clearly demonstrates the effectiveness of the proposed noise label learning framework. More qualitative segmentation results are shown in Figure~\ref{vis_coco_segmentation}. For example, in the second row, the segmentation network trained with our generated pseudo labels can partially mark the regions belonging to class of ``backpack'' (marked as orange), which are improperly labeled in ground truth annotations. Similarly, in the third row, our proposed  method can also successfully mark the class of ``dining table'' (marked as green) mislabeled by ground truth. These results clearly demonstrate the effectiveness of our method.

\section{Conclusion}
In this paper, we present a novel perspective for semi-supervised semantic segmentation and formulate the task as a problem of learning with pixel-level label noise. We design a label noise detection and correction framework, which transforms an image to a graph-structured representation and leverage GAT to address the noisy labels in pixel-level task. The experiments show that our method outperforms previous methods and achieves the state-of-the-art performance on PASCAL VOC 2012, PASCAL-Context and MS-COCO datasets in the semi-supervised setting.

\bibliographystyle{ieeetr}
\bibliography{egbib}
\end{document}